\documentclass{article}

\usepackage[nonatbib, final]{neurips_2021}

\usepackage{customization}
\usepackage{xcolor}

\usepackage[utf8]{inputenc} 
\usepackage[T1]{fontenc}    
\usepackage{url}            
\usepackage{amsfonts}       
\usepackage{microtype}      

\usepackage[toc,page]{appendix}
\newcommand{\stoptocwriting}{%
  \addtocontents{toc}{\protect\setcounter{tocdepth}{-5}}}
\newcommand{\resumetocwriting}{%
  \addtocontents{toc}{\protect\setcounter{tocdepth}{\arabic{tocdepth}}}}

\title{Adversarial Attacks on Black Box Video Classifiers: Leveraging the Power of Geometric Transformations}

\author{%
  Shasha Li\thanks{Equal contribution. Corresponding author: Shasha Li (\texttt{sli057@ucr.edu})}, Abhishek Aich$^*$, Shitong Zhu, M. Salman Asif, Chengyu Song,\\ \textbf{Amit K. Roy-Chowdhury, Srikanth V. Krishnamurthy} \\
  University of California, Riverside, CA, USA
}

\begin{document}

\maketitle

\begin{abstract}
When compared to the image classification models, black-box adversarial attacks against video classification models have been largely understudied. This could be possible because, with video, the temporal dimension poses significant additional challenges in gradient estimation. Query-efficient black-box attacks rely on effectively estimated gradients towards maximizing the probability of misclassifying the target video. In this work, we demonstrate that such effective gradients can be searched for by parameterizing the temporal structure of the search space with geometric transformations. Specifically, we design a novel iterative algorithm Geometric TRAnsformed Perturbations (\geotrap), for attacking video classification models. \geotrap employs standard geometric transformation operations to reduce the search space for effective gradients into searching for a small group of parameters that define these operations. This group of parameters describes the geometric progression of gradients, resulting in a reduced and structured search space. Our algorithm inherently leads to successful perturbations with surprisingly few queries. For example, adversarial examples generated from \geotrap have better attack success rates with $\sim$ 73.55\% fewer queries compared to the state-of-the-art method for video adversarial attacks on the widely used Jester dataset. Overall, our algorithm exposes vulnerabilities of diverse video classification models and achieves new state-of-the-art results under black-box settings on two large datasets. Code is available here: \url{https://github.com/sli057/Geo-TRAP}
\end{abstract}
\captionsetup[figure]{list=no}
\captionsetup[table]{list=no}
\stoptocwriting
\section{Introduction}
\label{sec:intro}

Adversarial attacks are designed to expose vulnerabilities of Deep Neural Networks (DNNs). With real-world applications of video classification based on DNNs emerging \cite{ji20133d, choi2018subcategory, aich2021spatio}, a key question that arises is ``\textit{what type of adversarial inputs can mislead, and thus render video classification networks vulnerable}?'' Designing such adversarial attacks not only helps expose security flaws of DNNs, but can also potentially stimulate the design of more robust video classification models.

Adversarial attacks against image classification models have been studied in both \textit{white-box}  \cite{szegedy2013intriguing,goodfellow2014explaining,carlini2017towards,song2018physical, xiao2018characterizing} and \textit{black-box} \cite{papernot2016transferability,papernot2017practical, bhagoji2018practical,chen2017zoo,ilyas2018prior} settings. In the white-box setting, an adversary has full access to the model under attack, including its parameters and training settings (hyper-parameters, training data, etc.) 
In the black-box setting, an adversary only has partial information about the victim model, such as the predicted labels of the model. In the case of video classification models, adversarial attacks in both white-box and black-box settings have garnered some interest  \cite{wei2019sparse,li2019stealthy,chen2021appending,pony2020over,lo2020multav, jiang2019black, wei2020heuristic, yan2020sparse, zhang2020motion}, although the body of work here is more limited than the case of image classification models.

A common black-box attack paradigm is query-based,
wherein the attacker can
send queries to the victim model 
to collect the corresponding predicted labels, and thereby estimate the gradients needed for curating the adversarial examples. Unlike static images, videos naturally include additional information from the temporal dimension. This high dimensionality (i.e., sequence of frames instead of one image) poses challenges to black-box adversarial attacks against video classification models; in particular, significantly more queries are typically needed for estimating the gradients for crafting adversarial samples~\cite{jiang2019black, yan2020sparse, wei2020heuristic,zhang2020motion}.
\cite{jiang2019black} reduces the number of queries by adding perturbations on the patch level instead of at the pixel level; \cite{yan2020sparse, wei2020heuristic} propose to add perturbations only on key pixels. \cite{zhang2020motion} considers the intrinsic differences between images and videos (i.e., the temporal dimension), and proposes to use the optical-flow of clean videos as the motion prior for adversarial video generation. Similar to \cite{zhang2020motion}, we also explicitly consider the temporal dimension of video. However,  rather than fixing the temporal search space using the motion prior of clean videos, we propose to parameterize the temporal structure of the space with  geometric transformations. This results in a better structured and reduced search space, which allows us to generate successful attacks with much fewer queries in black-box settings than the state-of-the-art methods, including \cite{zhang2020motion}.

\paragraph{Contributions.} In this paper, we propose a novel query-efficient black-box attack algorithm against video classification models. Due to the extra temporal dimension, generating video perturbations by searching for effective gradients remains a challenging task given the exceedingly large search space. 
These gradients are estimated by searching for `directions' that maximize the probability of the victim model mis-classifying the crafted inputs. Our approach drastically reduces this large search space by defining this space with a small set of parameters that  
describe the geometric progression of gradients in the temporal dimension, resulting in a reduced and temporally structured search space. Conceptually, this parameterization of the temporal structure of the search space is performed using geometric transformations (e.g. affine transformations). 
We refer to our algorithm as Geometrically TRAnsformed Perturbations, or \geotrap. Despite this surprisingly simple strategy, \geotrap outperforms existing black-box video adversarial attack methods by significant margins ($\sim$ 1.8\% improvement in attack success rate with $\sim$ 73.55\% {\em fewer queries} for targeted attacks in comparison to the state-of-the-art \cite{zhang2020motion} on the Jester dataset \cite{materzynska2019jester}).



\section{Related Works}
\label{sec:baselines}
\renewcommand{\arraystretch}{1.2}
\begin{table}
\centering
\caption{\textbf{Comparison with state-of-the-art.} \geotrap, compared to current black-box attack methods for videos, doesn't train a different network to craft perturbations, and parameterizes the temporal dimension of videos in searching for effective perturbation directions.} 
\scriptsize{
\begin{tabular}{c|c|c|c}
\hline
\rowcolor{black!10}
\multirow{2}{1.5cm}{\centering \textbf{Methods}} & 
\multirow{2}{3cm}{\centering \textbf{WITHOUT training a ``perturbation'' network }} & 
\multirow{2}{3cm}{\centering \textbf{CONSIDER temporal dimension?}} &
\multirow{2}{3cm}{\centering \textbf{PARAMETERIZE temporal dimension?}}\\[1em]
\hline
\PatchAttack  & \ccross & \ccross  & \ccross\\
\hline
\HeuristicAttack & \ccheck& \ccross  & \ccross\\
\hline
\SparseAttack & \ccross & \ccross  & \ccross\\
\hline
\MotionSamplerAttack & \ccheck & \ccheck  & \ccross\\
\hline
\geotrap~{(\textbf{Ours})} & \ccheck & \ccheck  & \ccheck \\
\hline
\end{tabular}}
\label{tab:related-works}
\vspace*{-\baselineskip}
\end{table}

In this section, we review different black-box adversarial attacks strategies, and categorize our proposed method with respect to state-of-the-art black-box attacks designed for video classifiers. 

In most real-world attacks, the adversary only has partial information about the victim models, such as the predicted labels. In such black-box settings, the adversary can first attack a local surrogate model and then transfer these attacks to the target victim model \cite{liu2016delving,huang2019enhancing}, formally called as \textit{transferability-based} black-box attack. Alternatively, they may estimate the adversarial gradient with zero-order optimization methods such as Finite Differences (FD) or Natural Evolution Strategies (NES) by querying the victim model \cite{chen2017zoo,ilyas2018black,ilyas2018prior}, which is called \textit{query-based} black-box attack. \geotrap falls under the category of query-based black-box attacks (designed for videos).

Whilst several white-box attacks have been proposed for video classification models \cite{wei2019sparse,li2019stealthy,chen2021appending,pony2020over,lo2020multav}, black-box video attacks are relatively under explored. 
\VBAD is the first to propose a black-box video attack framework which uses a hybrid attack strategy of first generating initial perturbations for each video frame by attacking a local image classifier, and then updating the perturbations by querying the victim model. Compared to \PatchAttack, \geotrap does not require training a local classifier. \PatchAttack crafts video perturbations by treating each frame as a separate image, but reduces the search space of the gradient estimation by morphing the perturbations in patches/partitions. However, its attack performance has been shown to be inferior to that of a more recent approach \cite{zhang2020motion} (discussed below). \HeuristicAttack uses a query-based attack strategy, and  reduces the search space by generating adversarial perturbations only on heuristically selected key frames and salient regions. \SparseAttack reduces the search space by adding perturbations only on key frames using a reinforcement learning based framework. \MotionSamplerAttack proposed a query-based attack strategy that utilizes a motion excited sampler to obtain \textit{motion-aware} perturbation prior by using the optical-flow of the clean video. This motion-aware prior reduces the search space for gradients resulting in fewer queries.
Similar to \cite{zhang2020motion} but different from \cite{jiang2019black, yan2020sparse, wei2020heuristic}, \geotrap explicitly considers the temporal dimension of video in order to search for effective gradients. However unlike \cite{zhang2020motion}, \geotrap does not fix the temporal structure of the search space using a \textit{pre-computed fixed} motion prior, but parameterize it with simple geometric transformations. These black-box video attack methods are summarized in Table \ref{tab:related-works}.

\section{Attacking via Geometrically TRAnsformed Perturbations (\geotrap)}
\label{sec:geotrap}

\paragraph{Notation.} 
We denote the tuple of a video clip and its corresponding label as $(\bm{x}, y)$, which represents a data-point in the distribution $\mathcal{X}$. Each video sample $\bm{x}\in \mathbb{R}^{T \times H \times W \times C}$ has $T$ frames of $H$ height, $W$ width, and $C$ channels. We denote the victim video classification model as $\bm{f}_{\bm{\theta}}:\mathcal{X}\rightarrow\mathcal{Y}$, where $\bm{\theta}$ represents the model's parameters learned from the training subset of $\mathcal{X}$, via a mapping to the label space $\mathcal{Y}$. We further assume $\mathcal{X}$ consists of videos from  $\vert\mathcal{Y}\vert = K$ categories. To make the perturbations imperceptible to humans, we impose the perturbation budget $\rho_{\max}$ with the $\Vert\cdot\Vert_p$ norm. Throughout this paper, we consider $\Vert\cdot\Vert_{\infty}$ norm following \cite{li2019stealthy, jiang2019black,zhang2020motion} (the method can be extended to $p=1, 2$ norms).
To constrain $\Vert\cdot\Vert_{\infty}$ of perturbation below a budget $\rho_{\max}$, we use the $\texttt{clip}(\cdot)$ function to keep the perturbation pixel value in [$-\rho_{\max}, \rho_{\max}$]. The function $\texttt{sign}(\cdot)$ extracts the sign of given input variable. The superscript $i$, throughout the paper, denotes the iteration $i$. The subscript $t$ denotes the frame index. 
For clarity, we represent vectors/tensors with the bold font and scalars with the regular font.

\paragraph{Problem Statement.} We consider the scenario of attacking a standard video classification model using a query-based paradigm under \textbf{black-box settings} (assuming no access to $\bm{\theta}$ nor the training subset of $\mathcal{X}$). Specifically, we aim to craft perturbed videos $\bm{x}_{\text{adv}}$ with imperceptible differences from $\bm{x}$, in order to alter the decision of the target model $\bm{f}_{\bm{\theta}}$ via multiple queries to guide the gradient estimation. This problem can be mathematically formulated as follows.
\begin{align}
\begin{aligned}\label{equ:opt}
   \underset{\bm{x}_{\text{adv}}}{\operatorname{argmin}}\quad \mathcal{L}\big{(}\bm{f}_{\bm{\theta}}(\bm{x}_{\text{adv}}),y\big{)}\quad\text{s.t.}\quad\lVert \bm{x}_{\text{adv}}-\bm{x} \rVert_\infty \leq \rho_{\max}
\end{aligned}
 \end{align}
$\mathcal{L}\big{(}\bm{f}_{\bm{\theta}}(\bm{x}_{\text{adv}}),y\big{)}$ is the objective function, capturing  the similarity between the classifier's output and the ground truth label $y$, and varies with different attack goals (targeted or untargeted). The challenge is to obtain $\bm{x}_{\text{adv}}$ with as few queries as possible by estimating gradient $\bm{g^{\star}}  = \nabla_{\bm{x}_\text{adv}} \mathcal{L}\big{(}\bm{f}_{\bm{\theta}}(\bm{x}_\text{adv}),y\big{)} $, which is unknown in the considered black-box setting.

\paragraph{Overview of \geotrap.} We propose a novel iterative video perturbation framework that follows the principle of the Basic Iterative Method \cite{kurakin2016adversarialphysical} in order to fool $\bm{f}_{\bm{\theta}}$ under $\Vert\cdot\Vert_\infty$ norm as follows.
\begin{align}
    \bm{x}^{(0)}_{\text{adv}} = \bm{x},\quad \bm{x}^{(i)}_{\text{adv}} = \texttt{clip}\big{(}\bm{x}^{(i-1)}_{\text{adv}}-h\,\texttt{sign}(\textcolor{black}{\bm{g}^{(i)}})\big{)} 
    \label{eq:update-step}
\end{align}
where $h$ is a hyperparameter and $\bm{g}^{(i)}$ is the gradient estimated by querying the black-box victim model at the $i^{th}$ iteration using our proposed \geotrap algorithm.  As shown in \eqref{eq:update-step}, effective perturbations rely on the guidance of the gradient $\bm{g}^{(i)}$. Therefore, efficiently estimating $\bm{g}^{(i)}$ is at the core of \geotrap for successfully subverting video classifiers. We execute the following two steps in each iteration \textcolor{black}{to estimate $\bm{g}^{(i)}$}.

\begin{enumerate}[leftmargin=*]
    \item For any input video $\bm{x}^{(i)}$, a random noise tensor $\bm{r}_{\text{frame}}\in\mathbb{R}^{H\times W\times C}$
    and a set of geometric transformation parameters $\bm{\Phi}_{\text{warp}} \in\mathbb{R}^{T\times D}$
    are chosen with each element sampled from a standard normal distribution. $D$ represents the number of parameters needed for the geometric transformation of a single frame (details are provided in Section \ref{sec:Trans-Wrap}). In this setup, our search space for estimating $\bm{g}^{(i)}$ consists of $\bm{r}_{\text{frame}}$ and $\bm{\Phi}_{\text{warp}}$. 
    
    \item We then warp $\bm{r}_{\text{frame}}$ with  $\bm{\Phi}_{\text{warp}}$ to get the candidate  direction $\bm{\pi}=[\bm{r}_1, \bm{r}_2, \cdots, \bm{r}_T]\in\mathbb{R}^{T\times H\times W\times C}$ (see Algorithm \ref{algo:motion_wrap} \textsc{Trans-Warp}). $\bm{\pi}$ is then employed to compute a gradient estimator $\bm{\Delta}$ 
    by querying the black-box victim model with a standard gradient estimation algorithm (see Algorithm \ref{algo:gradient_estimation} \textsc{Grad-Est}). \textcolor{black}{The gradient estimator $\bm{\Delta}$ is then used to update $\bm{g}^{(i)}$}.
    
\end{enumerate}
    
\begin{figure}[t]
\centering
\includegraphics[height=18em]{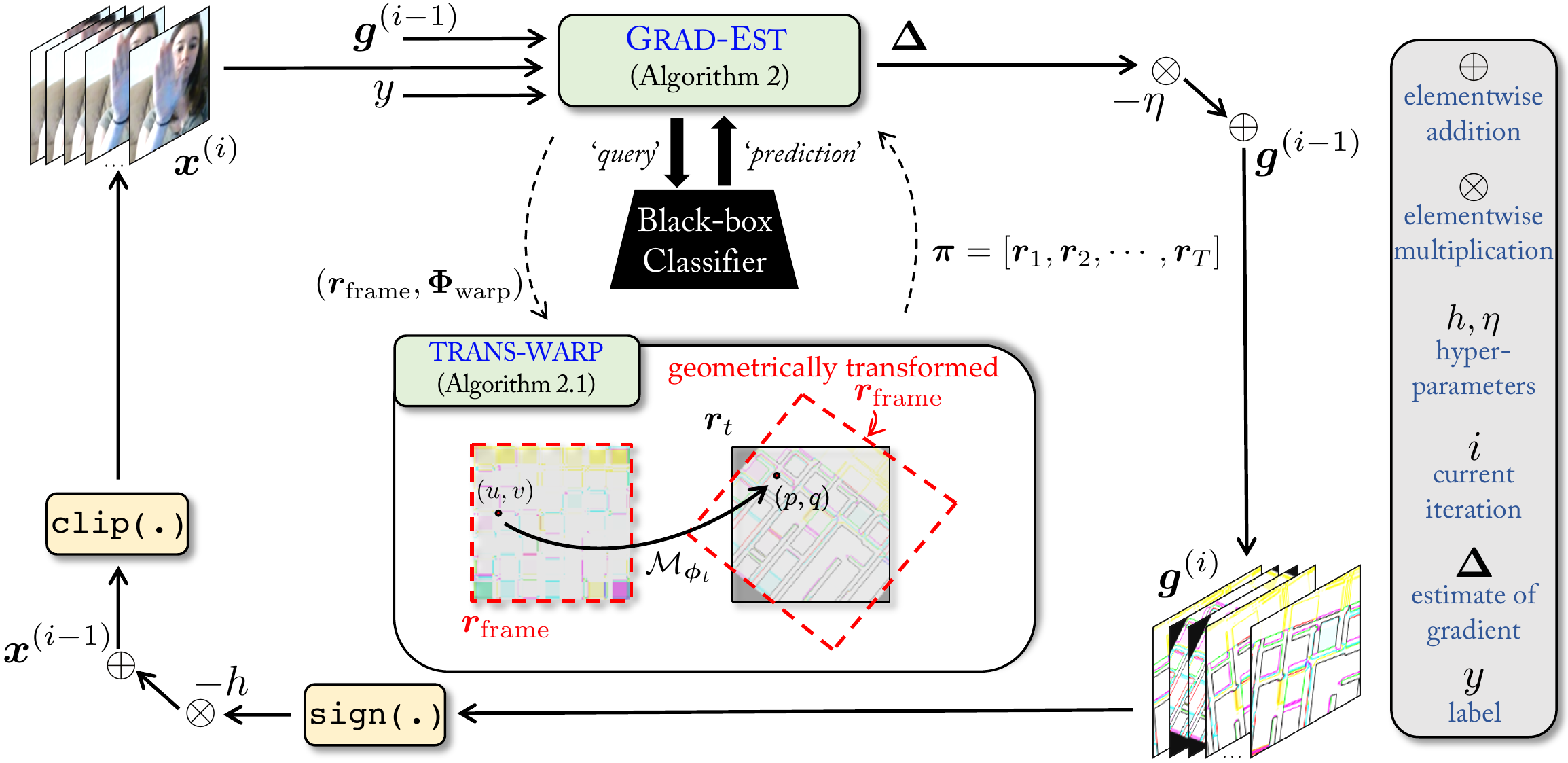}
\caption{\textbf{Overview of \geotrap.} \geotrap is a black-box attack algorithm guided by the key observation that strong gradients $\bm{g}^{(i)}$ can be computed by finding better gradient search direction candidates $\bm{\pi}$.
We propose to search each frame of the directions $\bm{r}_t$ by warping a randomly sampled $\bm{r}_{\text{frame}}$ using a geometric transformation $\mathcal{M}_{\bm{\phi}_{t}}$; different $\bm{r}_t$ in $\bm{\pi}$ are warped by the same $\bm{r}_{\text{frame}}$, thus have geometric progression among frames.  }
\label{fig:framework}
\vspace*{-\baselineskip}
\end{figure}

The overall attack strategy is summarized in Algorithm \ref{algo:adversairal_example_genration} and pictorially illustrated in Figure \ref{fig:framework}.
Since in Step 2 above, the gradient estimation (i.e., \textsc{Grad-Est}) procedure includes the geometric transformation strategy (i.e., \textsc{Trans-Warp}), we will next describe \textsc{Grad-Est} and then move on to \textsc{Trans-Warp}. For the simplicity of exposition, we drop the superscript $i$ and shorten the loss function to $\mathcal{L}(\bm{f}_{\bm{\theta}}(\bm{x}_{\text{adv}}),y)$ to $\mathcal{L}$ (as the model parameters $\bm{\theta}$ remain unchanged) in rest of this section.  

\subsection{\geotrap Gradient Estimation (\textsc{Grad-Est})}
Let $\bm{g}^\star = \nabla_{\bm{x}} \mathcal{L}$ be the ideal value of the gradient of $\mathcal{L}$ at $\bm{x}$, required to create $\bm{x}_{\text{adv}}$ in \eqref{eq:update-step}. 
To find an efficient estimator $\bm{g}$ for $\bm{g}^\star$, a (new) surrogate loss $\ell(\bm{g}) = - \langle \bm{g^\star}, \bm{g} \rangle$ is defined such that the estimator $\bm{g}$ has a sufficiently large inner product with the actual gradient $\bm{g^\star}$ \textcolor{black}{($\bm{g}$ is normalized to a unit vector; we ignore the normalization operation for ease of explanation)}. The loss function definition and the  algorithm to estimate $\bm{g}$ follow \cite{ilyas2018prior}.


As $\bm{g}^\star$ is unknown in the black-box setting, this surrogate loss function can be estimated as 
\begin{equation}
\label{equ:inner_product_est}
    \ell(\bm{g}) = -\langle \bm{g^\star}, \bm{g}\rangle  = - \langle  \nabla_{\bm{x}} \mathcal{L}, \bm{g}\rangle  \approx - \dfrac{\mathcal{L}(\bm{x}+\epsilon \bm{g} , y) - \mathcal{L}(\bm{x},y)}{\epsilon}. 
\end{equation}
To iteratively estimate $\bm{g}$, we need to, in turn, estimate the gradient of $\ell(\bm{g})$, i.e.,  $\bm{\Delta}=\nabla_{\bm{g}}\ell(\bm{g})$. With antithetic sampling \cite{ren2019adaptive}, 
$\bm{\Delta}$ can be estimated as 
\begin{equation}
\label{equ:gradient_est}
    \bm{\Delta} = \dfrac{\ell(\bm{g}+\delta \bm{\pi}) - \ell( \bm{g}-\delta \bm{\pi}) }{\delta}\bm{\pi}, 
\end{equation}
where $\delta$ is a small number adjusting the magnitude of the loss variation and $\bm{\pi}\in\mathbb{R}^{T\times H\times W\times C}$ is a random candidate direction. 
\textcolor{black}{Our core contribution lies in the fact that instead of randomly sampling $\bm{\pi}$ in the search space~\cite{ilyas2018prior}, we reduce the search dimensionality by warping a randomly sampled tensor $\bm{r}_{\text{frame}}\in\mathbb{R}^{H\times W\times C}$ with another randomly sampled geometric (e.g., affine) transformation parameter tensor 
$\bm{\Phi}_{\text{warp}} \in \mathbb{R}^{T\times D}$ to get $\bm{\pi}$. The search space is then reduced from $T \times H \times W \times C$ to ($H \times W \times C) + (T\times D$) and $D$ is a relatively small number, $D \ll H \times W \times C$.} With $\bm{w}_1= \bm{g}+\delta \bm{\pi} $ and $\bm{w}_2= \bm{g}-\delta \bm{\pi}$ and combining \eqref{equ:inner_product_est} with \eqref{equ:gradient_est}, we get
\begin{equation}
    \bm{\Delta} = \dfrac{\mathcal{L}(\bm{x}+\epsilon \bm{w}_2, y) - \mathcal{L}(\bm{x}+\epsilon \bm{w}_1, y)}{\epsilon \delta}\bm{\pi}. 
\end{equation}
Note that by querying the victim model $\bm{f}_{\bm{\theta}}$ with $\bm{x}+\epsilon\bm{w}_1$, we are able to retrieve the value of $\mathcal{L}(\bm{x}+\epsilon \bm{w}_1, y)$; similarly we can obtain the value of $\mathcal{L}(\bm{x}+\epsilon \bm{w}_2, y)$ (\textcolor{black}{$\mathcal{L}(\cdot)$ is defined following \cite{pony2020over}}.). 
\normalem
\begin{algorithm}[t]
	\selectfont
	\SetAlgoLined
	\SetKwInOut{Input}{Input}\SetKwInOut{Output}{Output}
	\SetKwFunction{GradEst}{\textcolor{black}{\textsc{Grad-Est}}}
	\SetKwFunction{sign}{sign}
	\SetKwFunction{clip}{clip}
	
	\Input{video $\bm{x}$, corresponding label $y$, step-size $\eta$ for updating the gradient, step-size $h$ for updating adversarial video.}
	\Output{adversarial video $\bm{x}_{\text{adv}}$}
	\textbf{Initialize}: $\bm{x}^{(0)} = \bm{x} $, $\bm{g}^{(0)}= \bm{0}, i=1 $ \\
	\While {$\operatorname{argmax} \big{(}\bm{f}_{\bm{\theta}}(\bm{x}^{(i)})\big{)} = y$}{
	$\bm{\Delta} = $ \GradEst{$\bm{x}^{(i-1)}, \bm{g}^{(i-1)}, y$}\quad\textmn{/* \tt{Gradient Estimation} */}\\
	$\bm{g}^{(i)} \leftarrow\bm{g}^{(i-1)} - \eta\bm{\Delta} $\\
	$\bm{x}^{(i)} \leftarrow $ \clip{$\bm{x}^{(i-1)} - h$\sign{$\bm{g}^{(i)}$}}\\
	$i\leftarrow i+1$ \\}
	\Return $\bm{x}_{\text{adv}}=\bm{x}^{(i)}$
	\caption{\geotrap: Query-based Iterative attack for Video Classifiers}
	\label{algo:adversairal_example_genration}
\end{algorithm}
In summary, we estimate $\bm{\Delta}$ with these two queries to the victim model.
The resulting algorithm for estimating gradient of $\nabla_{\bm{g}}\ell$ or $\bm{\Delta}$ for consequently estimating $\bm{g}$ is shown in Algorithm  \ref{algo:gradient_estimation}. Eventually at every iteration, we use $\bm{\Delta}$ to update $\bm{g}$ by applying a one-step gradient descent as $\bm{g} \leftarrow \bm{g} - \eta \bm{\Delta}$, where $\eta$ is a hyperparameter to update $\bm{g}$. This updated $\bm{g}$ is later used to obtain $\bm{x}_{\text{adv}}$ using \eqref{eq:update-step}.

\subsection{Noise Warping using Geometric Transformation (\textsc{Trans-Warp})}
\label{sec:Trans-Wrap}
To tackle the challenge of the high-dimensionality of the search space, we propose to parameterize the search space with a single random noise tensor $\bm{r}_{\text{frame}} \in \mathbb{R}^{ H \times W \times C}$ and a sequence of geometric transformations $\bm{\Phi}_{\text{warp}} \in \mathbb{R}^{T \times D}$. 
Apart from the reduction of the search space of gradient estimation, our geometric transformation provides a temporal structure to $\bm{\pi}$, which we discuss next. 

At every iteration, $\bm{\pi}=[\bm{r}_1,  \bm{r}_2, \dots, \bm{r}_T]$ represents the candidate  direction for \textcolor{black}{$\bm{\Delta}$}. 
These directions $\bm{r}_t\in\mathbb{R}^{H\times W\times C}$ are used to compute $\bm{\Delta}$ in order to update gradient $\bm{g}$.
To obtain $\bm{\pi}$, we use a sequence of transformation vectors $\bm{\Phi}_{\text{warp}}= [\bm{\phi}_1, \bm{\phi}_2, \dots, \bm{\phi}_T]$ where $\bm{\phi}_t \in \mathbb{R}^{D}$. The dimensionality $D$, chosen by the attacker, can vary depending on the transformation type that is populated from $\bm{\phi}_t$, \textit{e.g.}, $D=6$ for affine transformation. 
We take affine transformation as an example to describe the warping process. We start by randomly sampling $\bm{r}_{\text{frame}}$ and the sequence of $\bm{\phi}_t$ along with initializing each element in the sequence of $\bm{r}_t$ with zero in \textbf{every} iteration. \textsc{Trans-Warp} then computes $\bm{r}_t$ by warping $\bm{r}_{\text{frame}}$ using the parameters in $\bm{\phi}_t = [\phi_{11}^t, \phi_{12}^t, \phi_{13}^t, \phi_{21}^t, \phi_{22}^t, \phi_{23}^t]\in\mathbb{R}^{6}$ of $\bm{\Phi}_{\text{warp}}\in\mathbb{R}^{T\times 6}$ as follows. For all $C$ channels, let $(p, q)$ and $(u, v)$ be the target and source coordinates in $\bm{r}_{t}$ and $\bm{r}_{\text{frame}}$, respectively. $\bm{r}_t$ (for all channels) is computed as
\begin{align}
    \bm{r}_{t}(p, q) \leftarrow \bm{r}_{\text{frame}}(u,v),\quad 1 \leq p, u \leq H, ~ 1 \leq q, v \leq W.
    \label{eq:compact-warp}
\end{align}
Location $(p, q)$ is computed using the affine transform matrix $\mathcal{M}_{\bm{\phi}_t}$ created with $\phi_t$ in homogeneous coordinates~\cite{Hartley2004} as shown below. $t$ is dropped for simplicity.
\begin{equation}
    \begin{pmatrix}
    p\\
    q\\
    1
    \end{pmatrix}
    = \mathcal{M}_{\bm{\phi}} 
    \begin{pmatrix}
    u\\
    v\\
    1
    \end{pmatrix} 
    = 
    \begin{bmatrix}
    \phi_{11} & \phi_{12} & \phi_{13}\\
    \phi_{21} & \phi_{22} & \phi_{23}\\
    0         & 0         & 1 
    \end{bmatrix}
    \begin{pmatrix}
    u\\
    v\\
    1
    \end{pmatrix} 
    \label{equ:affine}
\end{equation}
We compactly denote this warping operation in \eqref{eq:compact-warp} and \eqref{equ:affine} 
with $\bm{r}_{t} =  \mathcal{T}(\bm{r}_\text{frame}, \bm{\phi}_t)$. 
Affine transformation allows translation, rotation, scaling, and skew to be applied to $\bm{r}_{\text{frame}}$ to get each $\bm{r}_t$. \textcolor{black}{Therefore, the sequence of $\bm{r}_t$ have affine geometric progression among its temporal dimension.} Other examples of geometric transformations may be more constrained, such as the similarity transformation $\mathcal{M}^{\text{S}}_{\bm{\phi}}$ (that allows translation, dilation (uniform scale) and rotation with $D=4$) and translation-dilation $\mathcal{M}^{\text{TD}}_{\bm{\phi}}$ (that allows translation and uniform dilation with $D=3$) as shown below.
\begin{equation}
\label{equ:dof}
    \small{
    [\phi_{11}, \phi_{12}, \phi_{13}, \phi_{23}]\rightarrow
    \mathcal{M}^{\text{S}}_{\bm{\phi}} = 
    \begin{bmatrix}
    \phi_{11} & \phi_{12} & \phi_{13}\\
    -\phi_{12} & \phi_{11} & \phi_{23}\\
    0 & 0 & 1 
    \end{bmatrix}
    ,[\phi_{11}, \phi_{13}, \phi_{23}]\rightarrow
        \mathcal{M}^{\text{TD}}_{\bm{\phi}} = 
    \begin{bmatrix}
    \phi_{11} & 0          & \phi_{13}\\
    0        & \phi_{11} & \phi_{23}\\
    0 & 0 & 1 
    \end{bmatrix}}
\end{equation}
\vspace*{-\baselineskip}

\begin{algorithm}[t]
	\selectfont
	\SetAlgoLined
	\SetKwInOut{Input}{Input}\SetKwInOut{Output}{Output}
	\SetKwFunction{WRAP}{\textcolor{black}{\textsc{Trans}-\textsc{Warp}}}
	
	\Input{video $\bm{x}^{(i)}$, label $y$,  gradient estimator $\bm{g}^{(i-1)}$, $\delta$ for loss variation, $\epsilon$ for approximation.}
	\Output{estimation of $\bm{\Delta}=\nabla_{\bm{g}} \ell(\bm{g})$}
	
	\textbf{Sample} $\bm{r_{\text{frame}}} \in \mathbb{R}^{H\times W\times C}$,
$\bm{\Phi}_{\text{warp}} \in \mathbb{R}^{T\times D}$  (each element from a normal distribution $\mathcal{N}(0,1)$) 
	\\
	$\bm{\pi} = $ \WRAP{$ \bm{r_{\text{frame}}}, \bm{\Phi}_{\text{warp}} $} \quad\textmn{/* \tt{Use Geometric Transformations} */}\\
	$\bm{w}_1 = \bm{g}^{(i-1)} + \delta \bm{\pi} $, $\bm{w}_2 = \bm{g}^{(i-1)} - \delta \bm{\pi} $\\
	$L_1 = \mathcal{L}(\bm{x}^{(i-1)} + \epsilon \bm{w_2}, y)$, $L_2 = \mathcal{L}(\bm{x}^{(i-1)} + \epsilon \bm{w_1}, y) $  \quad\textmn{/* \tt{Query victim model twice} */}\\
	$\bm{\Delta} = (L_{2} - L_{1})\sfrac{\bm{\pi}}{\epsilon \delta}$ \\
	\Return $\bm{\Delta}$
	\caption{\small \textsc{Grad}-\textsc{Est}($\bm{x}^{(i-1)}, \bm{g}^{(i-1)}\in\mathbb{R}^{T\times H\times W\times C}, y$)$\rightarrow$ Estimate $\bm{\Delta} = \nabla_{\bm{g}} \ell(\bm{g})\in\mathbb{R}^{T\times H\times W\times C}$}
	\label{algo:gradient_estimation}
\end{algorithm}
\section{What Makes \geotrap Effective?}
\label{sec:why}
\begin{figure}[htbp]
    \vspace*{-\baselineskip}
	\begin{subfigure}[t]{0.49\textwidth}
		\centering
		\includegraphics[width=0.75\textwidth]{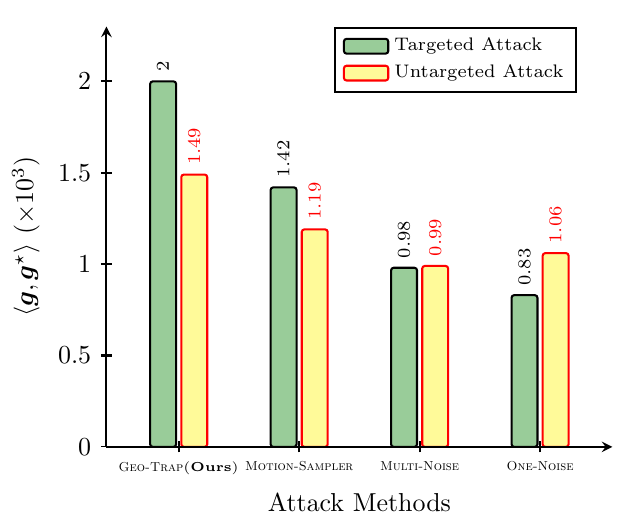}
		\caption{}
		\label{fig:gradient_curves}
	\end{subfigure}%
	\begin{subfigure}[t]{0.49\textwidth}
		\centering
		\includegraphics[width=0.7\textwidth]{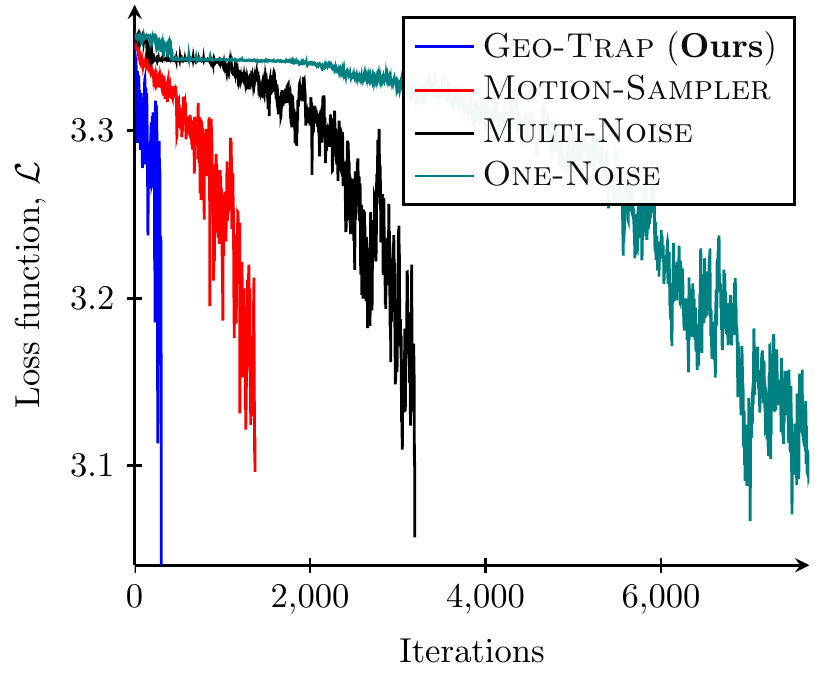}
		\caption{}
		\label{fig:loss_curves}
	\end{subfigure}
	\caption{\textbf{Gradient Analysis of \geotrap.} \textbf{(a)} \geotrap's high query-efficiency is a direct implication of good quality gradient estimation (for both targeted and untargeted attack), shown here with higher cosine similarity with $\bm{g}^\star$ compared to other methods. \textbf{(b)} Better quality of estimated gradients by \geotrap results in a successful attack with fewer queries compared to other attacks.}
	\label{fig:expain}
	\vspace*{-\baselineskip}
\end{figure}
Potent iterative algorithms should rely on few queries for crafting successful perturbations for time efficiency. To minimize the number of queries, iterative algorithms need to find strong gradients in their early iterations. As discussed earlier, videos inherently incur a larger search space due to the temporal dimension and thus, pose challenges in searching for effective gradients. 
In this section, we provide empirical evidence to show that by parameterizing the temporal dimension, \geotrap finds better gradients, in general, than previous works. We use three baselines in this analysis.

\renewcommand{\thealgocf}{2.\arabic{algocf}}
\setcounter{algocf}{0}
\begin{algorithm}[t]
	\selectfont
	\SetAlgoLined
	\SetKwInOut{Input}{Input}\SetKwInOut{Output}{Output}
	
	\Input{noise tensor $\bm{r}_{\text{frame}}$, warp tensors $\bm{\Phi}_{\text{warp}}$, transformation operation $\mathcal{T}_{\bm{\phi}}(\cdot)$.}
	\Output{candidate directions $\bm{\pi} = [\bm{r}_1, \bm{r}_2, \cdots, \bm{r}_T]$.}
	\textbf{Initialize} $\bm{\pi}=\emptyset$ \\
	\For{$t=[1, 2, \cdots, T]$}{
	$\bm{\phi}_t = \bm{\Phi}_{\text{warp}}[t]$\\
	$\bm{r}_t = \mathcal{T}(\bm{r}_{\text{frame}}, \phit)$\quad\textmn{/* \tt{Warping Operation} */} \\
	$\bm{\pi} \gets \text{append } \bm{r}_t $
	}
	\Return $\bm{\pi}$
	\caption{\small\textsc{Trans-Warp}($\bm{r}_{\text{frame}}\in\mathbb{R}^{H\times W\times C}$, $\bm{\Phi}_{\text{warp}}\in\mathbb{R}^{T\times D}$)$\rightarrow$ Estimate $\bm{\pi}\in\mathbb{R}^{T\times H\times W\times C}$}
	\label{algo:motion_wrap}
\end{algorithm}
\squishlist
\item \MultiNoiseAttack which computes search directions $\bm{r}_t$ separately for each frame by sampling each element of $\bm{r}_t$ from a standard normal distribution,
resulting in a search space dimension of $T\times H\times W\times C$. It does not explicitly consider the temporal dimension;  {\color{black} temporal progression in any arbitrary direction is possible between a sequence of perturbation frames}.  
\item \OneNoiseAttack which computes $\bm{r}_1$ by sampling each element from a standard normal distribution
and applies the same $\bm{r}_1$ across all $\bm{r}_t (t = 1, 2, \cdots, T)$. \OneNoiseAttack reduces the search space  but completely ignores the temporal dimension when generating the perturbation. 
\item \MotionSamplerAttack which uses the optical flow of the original video $\bm{x}$ to warp $\bm{r}_{\text{frame}}$ to get each $\bm{r}_t$. It reduces the search space by using the motion prior of $\bm{x}$ as the temporal progression between perturbation frames. In contrast, rather than fixing the temporal search space using a motion prior, \geotrap parameterizes the temporal structure of the space with $\bm{\Phi}_{\text{warp}}$.
\squishend

We measure the gradient estimation quality by calculating the cosine similarity between the ground truth $\bm{g}^\star$ and the estimated gradient $\bm{g}$ following \cite{jiang2019black} for the aforementioned baselines. For each attack, we average over 1000 randomly selected videos with their cosine similarity values in the first attack iteration. We choose the first iteration because the initial $\bm{g}^\star$ is the same for the different attack methods, ensuring a fair comparison. As shown in Figure \ref{fig:gradient_curves}, our proposed method for estimating the gradients, yields $\bm{g}$ of the best quality for both untargeted and targeted attacks among all evaluated approaches. This leads to faster loss convergence / few queries as shown in Figure \ref{fig:loss_curves}. We validate such trends with different loss functions and more datasets in the Supplementary Material.

The empirical results validate that by carefully considering the temporal dimension and parameterizing the temporal structure of the gradient search space with geometric transformations, \geotrap finds better gradients. \textcolor{black}{\geotrap and \MotionSamplerAttack are better than the other two; the reason could be that temporally structured perturbations are more likely to disrupt the motion context of videos. However, the gradients estimated by \MotionSamplerAttack are not as effective as our proposed approach; the reason could be that the motion-prior of the clean video does not necessarily represent the temporal behavior of effective video perturbations.}
By allowing flexibility of the temporal progression  while maintaining only a minimally sufficient space through its geometric parameterization, \geotrap generates effective temporally structured perturbations. \textcolor{black}{Note that one could use other, potentially better ways to parameterize the temporal progression of the video perturbation; this is regarded as future works.}
\vspace*{-1\baselineskip}

\section{Experiments}
\label{sec:experiments}
\paragraph{Datasets.} Following previous work like \cite{li2019stealthy}, we use the human action recognition dataset UCF-101 \cite{soomro2012ucf101} and the hand gesture recognition dataset 20BN-JESTER (Jester) \cite{materzynska2019jester} to validate our attacks. \textit{UCF-101} includes 13320 videos from 101 human action categories (e.g., applying lipstick, biking, blow drying hair, cutting in the kitchen). Given the diversity it provides, we consider the dataset to validate the feasibility of our attacks  on \textit{coarse-grained} actions. \textit{Jester}, on the other hand, includes hand gesture videos that are recorded by crowd-sourced workers performing 27 kinds of gestures (e.g., sliding hand left, sliding two fingers left, zooming in with full hand, zooming out with full hand). The appearance of different hand gestures is similar; it is the motion information that matters in the video classification. We use this dataset to validate our attack with regard to \textit{fine-grained} actions. 

\paragraph{Baselines.} Among the four state-of-the-art black-box video attack methods~\cite{jiang2019black, yan2020sparse, wei2020heuristic,zhang2020motion} described in Section~\ref{sec:baselines}, we use \cite{zhang2020motion, yan2020sparse} as baselines for following reasons. Our first baseline is \MotionSamplerAttack, which has been shown to outperform \PatchAttack, \textsc{One-Noise} and \textsc{Multi-Noise} attacks (introduced in Section.~\ref{sec:why}). Our second baseline is \HeuristicAttack. We note that \SparseAttack is not included in our analysis as we couldn't replicate their results.

\paragraph{Attack Settings.} 
\textcolor{black}{We consider four state-of-the-art video classification models representing diverse methodologies of learning from videos, i.e., C3D \cite{tran2015learning}, SlowFast \cite{feichtenhofer2019slowfast}, TPN \cite{yang2020temporal} and I3D \cite{,carreira2017quo}, as our black-box victim models to attack. More details about the four video models are  provided in the Supplementary Materials.}
For UCF-101, we randomly select one video from each category following the setting in \cite{jiang2019black,zhang2020motion}. For Jester, since the number of categories is small, we randomly select four videos from each category. All attacked videos are correctly classified by the black-box model. For targeted attack, a random target class is chosen for each video.  
The maximum noise value $\rho_{\max}$ is 10 pixel values (out of 255) following \cite{moosavi2017universal,mopuri2018nag,li2019stealthy}. We provide more results for different $\rho_{\max}$ in Supplementary Material. Note that since the perturbation generated by \HeuristicAttack is sparse and thus more imperceptible, we do not impose a perturbation budget on it. We set the maximum query limit to $Q=60,000$ for untargeted attack and $Q=200,000$ for targeted attack. \textcolor{black}{The other hyper-parameters, i.e., $\epsilon$, $\delta$, $\eta$, and $h$ take the same values as mentioned in \cite{zhang2020motion}.} Unless otherwise specified, a translation-dilation transformation (with $D=3$) is used for our attack method. 

\paragraph{Metrics.} Following \cite{jiang2019black,zhang2020motion}, we evaluate  
\geotrap, in terms of \textit{(a)} Success Rate (SR), i.e., the total success rate in attacking within query and perturbation budgets; and \textit{(b)} Average Number of Queries (ANQ) i.e., the average  total queries from attacks for all videos (including failed ones).

\begin{figure}[t]
	\centering
	\includegraphics[height=0.6\columnwidth]{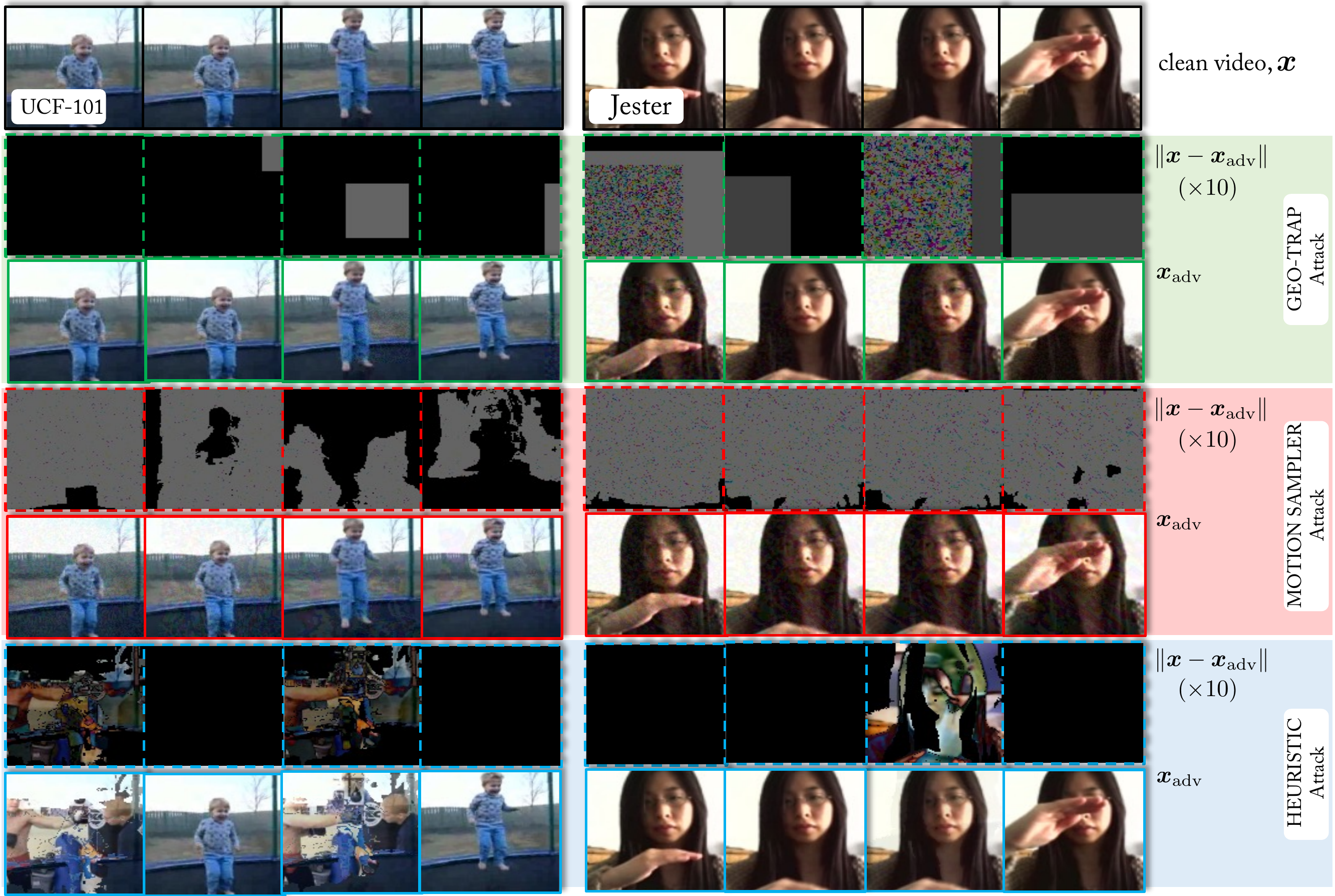}
	\vspace*{\baselineskip}
	\caption{\textbf{Visualization of Perturbations and Perturbed Video.} We visualize the generated perturbations and perturbed video for \geotrap and other baselines for UCF-101 (\textit{left}) and Jester (\textit{right}) datasets for untargeted attack against SlowFast classifier with $\rho_{\max}=\nicefrac{10}{255}$.} 
	\label{fig:visualization}
	\vspace*{-1.5\baselineskip}
\end{figure}

\subsection{Comparison to State-of-the-Art}

\paragraph{Untargeted Attack.} We report the untargeted attack performance of our attack method and the baseline methods in Table \ref{tab:untargeted_attack}. We observe that, in general, \geotrap requires fewer average number of queries when attacking different black-box victim models: on average over 45 \% fewer queries than \MotionSamplerAttack.  At the same time, \geotrap yields higher attack success rates: on average about  6\% higher than \HeuristicAttack.
When attacking SlowFast model on the Jester dataset, \geotrap achieves 100\% successful rate with only 521 queries while the baseline methods need at least 1906 queries. 
We also observe that the TPN model is more robust towards black-box attacks compared with the other three video recognition models. 

\paragraph{Visualization.} We show two visualizations of adversarial frames on Jester and UCF-101 in Fig.~\ref{fig:visualization}. We observe that the generated adversarial frames have little difference from the clean ones but can lead to a failed classification. Also, our attack method could lead to \textit{sparse} perturbations in the {spatial} and temporal dimension as the perturbations are sometimes zoomed out (thus get very small), and sometimes are translated out of the sight with choice of geometric transformation. More examples are in the Supplementary Material.
\renewcommand{\arraystretch}{1.2}
\begin{table}[b]
\centering
\vspace*{-1\baselineskip}
\caption{\textbf{Untargeted Attacks}. \geotrap demonstrates highly successful untargeted attacks (high Success Rate (SR)) 
with fewer queries (low Average Number of Queries (ANQ)) 
}

\label{tab:untargeted_attack}
\resizebox{\columnwidth}{!}{
\begin{tabular}{c|c|cc|cc|cc|cc}
\hline
\rowcolor{black!10}
 &  & \multicolumn{8}{c}{\textbf{Black-box Video Classifiers}}                                                 \\
 \cline{3-10}
\rowcolor{black!10}
 &  & \multicolumn{2}{c|}{\textbf{C3D}}  & \multicolumn{2}{c|}{\textbf{SlowFast}}                                           & \multicolumn{2}{c|}{\textbf{TPN}}                                                 & \multicolumn{2}{c}{\textbf{I3D}}                                                 \\
 \cline{3-10}
\rowcolor{black!10} \multirow{-3}{*}{\textbf{Datasets}}
                         &       \multirow{-3}{*}{\textbf{Methods}}           & ANQ ($\downarrow$)                     & SR ($\uparrow$)                                 & ANQ ($\downarrow$)                             & SR ($\uparrow$)                                 & ANQ ($\downarrow$)                               & SR ($\uparrow$)                                  & ANQ ($\downarrow$)                              & SR ($\uparrow$)                                  \\ \hline
\multirow{3}{*}{Jester}  & \HeuristicAttack        & 4699                     & 99.0\%                              & 3572                             & 98.1\%                              & 4679                              & 82.0\%                              & 4248                              & 98.1\%                              \\ \cline{2-10} 
                         & \MotionSamplerAttack    & 4549                     & 99.0\%                              & 1906                             & 100\%                              & 6269                              & 91.3\%                              & 3029                              & 99.4\%                              \\ \cline{2-10} 
                         & \geotrap (\textbf{Ours})                    & \textbf{1602}            & \textbf{100\%}                     & \textbf{521}                     & \textbf{100\%}                      & \textbf{3315}                     & \textbf{92.4\%}                     & \textbf{1599}                     & \textbf{100\%}                      \\ \hline
\multirow{3}{*}{UCF-101}     & \HeuristicAttack        & \textbf{5206} & 70.2\%          & 3507         & 87.2\%          & \textbf{6539}         &  71.8\%          &  6949          &  84.7\%          \\ \cline{2-10} 
                         & \MotionSamplerAttack    & 14336                     &  81.6\%          &  4673         &  97.2\%          &  20369          &  75.8\%          &  7400          &  94.4\%          \\ \cline{2-10} 
                         & \geotrap (\textbf{Ours})                    & {11490}            &  \textbf{86.2\%} &  \textbf{1547} &  \textbf{98.8\%} &  {17716} &  \textbf{76.1\%} &  \textbf{4887} &  \textbf{97.4\%} \\ \hline
\end{tabular}
}
\vspace*{-\baselineskip}
\end{table}

\paragraph{Targeted Attack.} We report the targeted attack performance of our  method and the baseline methods in Table.~\ref{tab:targeted_attack}. We observe that in some cases, \HeuristicAttack requires fewer number of queries than \geotrap, but its attack success rates are pretty low in those cases. For example, when attacking the TPN model on the Jester dataset, although \HeuristicAttack requires only 12k average number of queries, its attack success rate is less than half of ours, 44.4\% v.s. 92.6\%. 
\geotrap consistently yields higher attack success rates, on average over  30\% higher than \HeuristicAttack and over 8\% higher than \MotionSamplerAttack. In addition, in most cases, \geotrap requires fewer average number of queries than the two baseline attacks, on average over 45 \% fewer queries than \MotionSamplerAttack. The targeted attack performance further validates the effectiveness of our method. 
\renewcommand{\arraystretch}{1.2}
\begin{table}[t]
\centering
\caption{\textbf{Targeted Attacks}. \geotrap demonstrates highly successful targeted attacks (high Success Rate (SR)) 
with fewer queries (low Average Number of Queries (ANQ)) 
}
\label{tab:targeted_attack}
\resizebox{\columnwidth}{!}{
\begin{tabular}{c|c|cc|cc|cc|cc}
\hline
\rowcolor{black!10}
 &  & \multicolumn{8}{c}{\textbf{Black-box Video Classifiers}}                                                 \\
 \cline{3-10}
\rowcolor{black!10}
 &  & \multicolumn{2}{c|}{\textbf{C3D}}  & \multicolumn{2}{c|}{\textbf{SlowFast}}                                           & \multicolumn{2}{c|}{\textbf{TPN}}                                                 & \multicolumn{2}{c}{\textbf{I3D}}                                                 \\
 \cline{3-10}
\rowcolor{black!10} \multirow{-3}{*}{\textbf{Datasets}}
                         &       \multirow{-3}{*}{\textbf{Methods}}           & ANQ ($\downarrow$)                     & SR ($\uparrow$)                                 & ANQ ($\downarrow$)                             & SR ($\uparrow$)                                 & ANQ ($\downarrow$)                               & SR ($\uparrow$)                                  & ANQ ($\downarrow$)                              & SR ($\uparrow$)                                  \\ \hline
\multirow{3}{*}{Jester}  & \HeuristicAttack &      15595    &    46.3\%     &      30768          &      98.1\%                               &          \textbf{12006}                          &        44.4\%             &       31088       &   77.8\%   \\ \cline{2-10} 
                         & \MotionSamplerAttack    & 26704                & 98.2\%                              & 33087                              & 100\%                               & 63721                              & 80.9\%                              & 39037                              & 90.7\%                              \\ \cline{2-10} 
                         &  \geotrap (\textbf{Ours})                    & \textbf{6198}        & \textbf{100\%}                      & \textbf{7788}                      & \textbf{100\%}                      & {41294}                     & \textbf{92.6\%}                     & \textbf{19542}                     & \textbf{98.2\%}                     \\ \hline
\multirow{3}{*}{UCF-101}     & \HeuristicAttack        & \textbf{26741}  &   29.0\%      &        {22152}        &   61.4\%               &       \textbf{71828}          &    36.4\%             &       {92244}          &      43.7\%    \\ \cline{2-10} 
                         & \MotionSamplerAttack    & 100467                &  71.1\%          &  57126          &  86.0\%          &  {151409} &  31.6\%          &  {96498} &  59.6\%          \\ \cline{2-10} 
                         &  \geotrap (\textbf{Ours})                    & {71820}       &  \textbf{85.8\%} &  \textbf{21878} &  \textbf{95.0\%} &  141629          &  \textbf{40.0\%} &  \textbf{76708}          &  \textbf{74.6\%} \\ \hline
\end{tabular}
}
\vspace*{-\baselineskip}
\end{table}
\subsection{Different Geometric Transformations in \textsc{Trans-warp}}
\begin{wrapfigure}[15]{R}{0.425\textwidth}
    \centering
    \vspace*{-1\baselineskip}
    \usepgfplotslibrary{groupplots}
\usetikzlibrary{decorations.pathmorphing}
\def\axisdefaultwidth{240pt}
\pgfplotsset{
  every axis/.append style = {thick},tick style = {thick,black},
  %
  /tikz/normal shift/.code 2 args = {%
    \pgftransformshift{%
        \pgfpointscale{#2}{\pgfplotspointouternormalvectorofticklabelaxis{#1}}%
    }%
  },%
  range3frame/.style = {
    tick align        = outside,
    scaled ticks      = false,
    enlargelimits     = false,
    ticklabel shift   = {10pt},
    axis lines*       = left,
    line cap          = round,
    clip              = false,
    xtick style       = {normal shift={x}{10pt}},
    ytick style       = {normal shift={y}{10pt}},
    ztick style       = {normal shift={z}{10pt}},
    x axis line style = {normal shift={x}{10pt}},
    y axis line style = {normal shift={y}{10pt}},
    z axis line style = {normal shift={z}{10pt}},
  }
}

\pgfplotsset{every axis legend/.append style={    
    at={(0.05,1.10)},
    anchor=north west,
    font = \scriptsize,
    legend columns=-1,
    /tikz/every even column/.append style={column sep=0.15cm}
        }}
\begin{tikzpicture}[scale=0.65, transform shape]
\begin{axis}[
legend cell align=left,
every axis plot post/.style={/pgf/number format/fixed},
ybar=2pt,
bar width=10pt,
ymin=0,
axis on top,
ymax=3500,
xtick=data,
axis lines = left,
ylabel=ANQ ($\downarrow$),
enlarge x limits=0.2,
symbolic x coords={C3D,SlowFast,TPN,I3D},
visualization depends on=rawy\as\rawy, 
after end axis/.code={ 
\draw [ultra thick, white] (rel axis cs:0,1.05) -- (rel axis cs:1,1.05);
    },
nodes near coords={\scriptsize{\pgfmathprintnumber{\rawy}}},
every node near coord/.append style={
                        anchor=west,
                        rotate=90
                },
axis lines*=left,
clip=false,
area legend
]
\addplot[fill=red!40, rounded corners=1pt] coordinates {(C3D,1602) (SlowFast,521) (TPN,1797) (I3D,1599)};
\addplot[fill=yellow!40, rounded corners=1pt] coordinates {(C3D,1621) (SlowFast,532)  (TPN,1873) (I3D,1629)};
\addplot[fill=blue!40, rounded corners=1pt] coordinates {(C3D,2716) (SlowFast,1057) (TPN,2787) (I3D,2679)};
\legend{Translation Dilation,Similarity,Affine};
\end{axis}

\end{tikzpicture}
    \caption{\textbf{Performance with different} $\mathcal{M}_{\bm{\phi}}$. \geotrap results in best performance when $\mathcal{M}_{\bm{\phi}}$ is set as translation-dilation operation.}
    \label{fig:dof}
\end{wrapfigure}
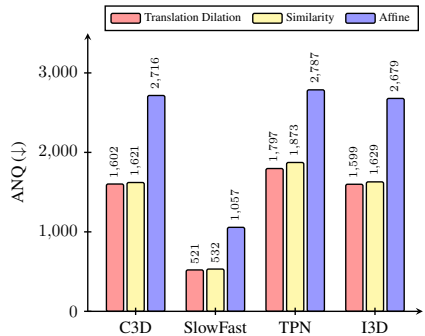
As discussed in Section \ref{sec:Trans-Wrap}, different kinds of geometric transformations could be used in the \textsc{Trans-Warp} function. 
In addition to the translation-dilation transformation ($\mathcal{M}^{\text{TD}}_{\bm{\phi}}$ in \eqref{equ:dof}, $D=3$) employed throughout the paper, we report the performance of \geotrap with two other different geometric transformations, i.e., similarity transformation ($\mathcal{M}^{\text{S}}_{\bm{\phi}}$ in \eqref{equ:dof}, $D=4$) and affine transformation ($\mathcal{M}_{\bm{\phi}}$ in \eqref{equ:affine}, $D=6$). Figure \ref{tab:dof} shows the untargeted attack performance on Jester with these different geometric transformations (more results are available in the Supplementary Material). We observe that the transformation with fewer degrees of freedom (DOF) ( translation-dilation transformation) tends to require fewer queries while having the same or higher attack success rates \textcolor{black}{(the attack success rates are available in the Supplementary Material). We believe that $D=3$ provides enough temporal flexibility to disrupt the motion context of the videos; additional degrees of freedom seemingly increase the search space unnecessarily, resulting in more queries.}

\section{Conclusion}
Black-box adversarial attacks on video classifiers is a challenging problem that has been largely understudied. In this work, we demonstrate that searching for effectual gradients in a reduced but structured search space for crafting perturbations leads to highly successful attacks with fewer queries compared to state-of-the-art attack strategies. In particular, we propose a novel iterative algorithm that employs Geometric transformations to parameterize and reduce the search space, for estimating gradients that maximize the probability of mis-classification of the perturbed video. This simple and novel strategy exposes the vulnerability of widely used video classifiers. For instance, \geotrap decreases average query numbers by 64.78\%, 72.66\% and 47.21\% to attack C3D, SlowFast, and I3D, respectively, for almost 100\% success rate in untargeted attacks.

\section{Broader Impact}
In this work, by leveraging geometric transformations for effective gradient estimations, we propose a highly query-efficient adversarial attack on video classification models which demonstrates state-of-the-art results. As more and more safety-critical systems (e.g., perceptual modules in autonomous vehicles) nowadays rely on video models, we are hopeful that our work, in addition to future research (including designing sophisticated video generative models as in \cite{aich2020non, 10.1145/3394171.3413686, Tulyakov:2018:MoCoGAN} for distribution-driven attacks \cite{naseer2019cross, poursaeed2018generative}), can eventually help build sufficiently robust video models to best avoid malicious sub-versions. On one hand, we believe our algorithm could allow further research in adversarial robustness and data augmentation strategies of deep vision models. It should also give a direction to researchers to design counter defense methodologies \cite{yin2019gat,wang2019convergence,li2020connecting, wang2021multi, yin2021exploiting}. On the other hand, it highlights a key drawback of different video classifiers which will allow adversaries to design more sophisticated attacks, both in white-box and black-box settings. Addressing such fallacies in designing deep neural networks is of utmost importance before introducing them in real-world scenarios. 

\paragraph{Acknowledgement.} The authors would like to thank Dr. Cliff Wang of US Army Research Office for his extensive comments and input on this work. This material is based upon work supported by the Defense Advanced Research Projects Agency (DARPA) under Agreement No. HR00112090096. Approved for public release; distribution is unlimited. 

\bibliographystyle{unsrtnat}
\bibliography{ref.bib}

\begin{thebibliography}{47}
\providecommand{\natexlab}[1]{#1}
\providecommand{\url}[1]{\texttt{#1}}
\expandafter\ifx\csname urlstyle\endcsname\relax
  \providecommand{\doi}[1]{doi: #1}\else
  \providecommand{\doi}{doi: \begingroup \urlstyle{rm}\Url}\fi

\bibitem[Ji et~al.(2013)Ji, Xu, Yang, and Yu]{ji20133d}
Shuiwang Ji, Wei Xu, Ming Yang, and Kai Yu.
\newblock 3d convolutional neural networks for automatic human action
  recognition, January~1 2013.
\newblock US Patent 8,345,984.

\bibitem[Choi et~al.(2018)Choi, Lin, Xiang, and Savarese]{choi2018subcategory}
Wongun Choi, Yuanqing Lin, Yu~Xiang, and Silvio Savarese.
\newblock Subcategory-aware convolutional neural networks for object detection,
  May~8 2018.
\newblock US Patent 9,965,719.

\bibitem[Aich et~al.(2021)Aich, Zheng, Karanam, Chen, Roy-Chowdhury, and
  Wu]{aich2021spatio}
Abhishek Aich, Meng Zheng, Srikrishna Karanam, Terrence Chen, Amit~K
  Roy-Chowdhury, and Ziyan Wu.
\newblock Spatio-temporal representation factorization for video-based person
  re-identification.
\newblock \emph{arXiv preprint arXiv:2107.11878}, 2021.

\bibitem[Szegedy et~al.(2013)Szegedy, Zaremba, Sutskever, Bruna, Erhan,
  Goodfellow, and Fergus]{szegedy2013intriguing}
Christian Szegedy, Wojciech Zaremba, Ilya Sutskever, Joan Bruna, Dumitru Erhan,
  Ian Goodfellow, and Rob Fergus.
\newblock {Intriguing Properties of Neural Networks}.
\newblock \emph{arXiv preprint arXiv:1312.6199}, 2013.

\bibitem[Goodfellow et~al.(2014)Goodfellow, Shlens, and
  Szegedy]{goodfellow2014explaining}
Ian~J Goodfellow, Jonathon Shlens, and Christian Szegedy.
\newblock {Explaining and Harnessing Adversarial Examples}.
\newblock \emph{arXiv preprint arXiv:1412.6572}, 2014.

\bibitem[Carlini and Wagner(2017)]{carlini2017towards}
Nicholas Carlini and David Wagner.
\newblock Towards evaluating the robustness of neural networks.
\newblock In \emph{2017 ieee symposium on security and privacy (sp)}, pages
  39--57. IEEE, 2017.

\bibitem[Song et~al.(2018)Song, Eykholt, Evtimov, Fernandes, Li, Rahmati,
  Tramer, Prakash, and Kohno]{song2018physical}
Dawn Song, Kevin Eykholt, Ivan Evtimov, Earlence Fernandes, Bo~Li, Amir
  Rahmati, Florian Tramer, Atul Prakash, and Tadayoshi Kohno.
\newblock Physical adversarial examples for object detectors.
\newblock In \emph{12th $\{$USENIX$\}$ Workshop on Offensive Technologies
  ($\{$WOOT$\}$ 18)}, 2018.

\bibitem[Xiao et~al.(2018)Xiao, Deng, Li, Yu, Liu, and
  Song]{xiao2018characterizing}
Chaowei Xiao, Ruizhi Deng, Bo~Li, Fisher Yu, Mingyan Liu, and Dawn Song.
\newblock Characterizing adversarial examples based on spatial consistency
  information for semantic segmentation.
\newblock In \emph{Proceedings of the European Conference on Computer Vision
  (ECCV)}, pages 217--234, 2018.

\bibitem[Papernot et~al.(2016)Papernot, McDaniel, and
  Goodfellow]{papernot2016transferability}
Nicolas Papernot, Patrick McDaniel, and Ian Goodfellow.
\newblock Transferability in machine learning: from phenomena to black-box
  attacks using adversarial samples.
\newblock \emph{arXiv preprint arXiv:1605.07277}, 2016.

\bibitem[Papernot et~al.(2017)Papernot, McDaniel, Goodfellow, Jha, Celik, and
  Swami]{papernot2017practical}
Nicolas Papernot, Patrick McDaniel, Ian Goodfellow, Somesh Jha, Z~Berkay Celik,
  and Ananthram Swami.
\newblock Practical black-box attacks against machine learning.
\newblock In \emph{Proceedings of the 2017 ACM on Asia conference on computer
  and communications security}, pages 506--519, 2017.

\bibitem[Bhagoji et~al.(2018)Bhagoji, He, Li, and Song]{bhagoji2018practical}
Arjun~Nitin Bhagoji, Warren He, Bo~Li, and Dawn Song.
\newblock Practical black-box attacks on deep neural networks using efficient
  query mechanisms.
\newblock In \emph{Proceedings of the European Conference on Computer Vision
  (ECCV)}, pages 154--169, 2018.

\bibitem[Chen et~al.(2017)Chen, Zhang, Sharma, Yi, and Hsieh]{chen2017zoo}
Pin-Yu Chen, Huan Zhang, Yash Sharma, Jinfeng Yi, and Cho-Jui Hsieh.
\newblock Zoo: Zeroth order optimization based black-box attacks to deep neural
  networks without training substitute models.
\newblock In \emph{Proceedings of the 10th ACM workshop on artificial
  intelligence and security}, pages 15--26, 2017.

\bibitem[Ilyas et~al.(2018{\natexlab{a}})Ilyas, Engstrom, and
  Madry]{ilyas2018prior}
Andrew Ilyas, Logan Engstrom, and Aleksander Madry.
\newblock Prior convictions: Black-box adversarial attacks with bandits and
  priors.
\newblock \emph{arXiv preprint arXiv:1807.07978}, 2018{\natexlab{a}}.

\bibitem[Wei et~al.(2019)Wei, Zhu, Yuan, and Su]{wei2019sparse}
Xingxing Wei, Jun Zhu, Sha Yuan, and Hang Su.
\newblock Sparse adversarial perturbations for videos.
\newblock In \emph{Proceedings of the AAAI Conference on Artificial
  Intelligence}, pages 8973--8980, 2019.

\bibitem[Li et~al.(2019)Li, Neupane, Paul, Song, Krishnamurthy, Roy-Chowdhury,
  and Swami]{li2019stealthy}
Shasha Li, Ajaya Neupane, Sujoy Paul, Chengyu Song, Srikanth~V Krishnamurthy,
  Amit~K Roy-Chowdhury, and Ananthram Swami.
\newblock Stealthy adversarial perturbations against real-time video
  classification systems.
\newblock In \emph{NDSS}, 2019.

\bibitem[Chen et~al.(2021)Chen, Xie, Pang, He, and Tian]{chen2021appending}
Zhikai Chen, Lingxi Xie, Shanmin Pang, Yong He, and Qi~Tian.
\newblock Appending adversarial frames for universal video attack.
\newblock In \emph{Proceedings of the IEEE/CVF Winter Conference on
  Applications of Computer Vision}, pages 3199--3208, 2021.

\bibitem[Pony et~al.(2020)Pony, Naeh, and Mannor]{pony2020over}
Roi Pony, Itay Naeh, and Shie Mannor.
\newblock Over-the-air adversarial flickering attacks against video recognition
  networks.
\newblock \emph{arXiv preprint arXiv:2002.05123}, 2020.

\bibitem[Lo and Patel(2020)]{lo2020multav}
Shao-Yuan Lo and Vishal~M Patel.
\newblock Multav: Multiplicative adversarial videos.
\newblock \emph{arXiv preprint arXiv:2009.08058}, 2020.

\bibitem[Jiang et~al.(2019)Jiang, Ma, Chen, Bailey, and Jiang]{jiang2019black}
Linxi Jiang, Xingjun Ma, Shaoxiang Chen, James Bailey, and Yu-Gang Jiang.
\newblock Black-box adversarial attacks on video recognition models.
\newblock In \emph{Proceedings of the 27th ACM International Conference on
  Multimedia}, pages 864--872, 2019.

\bibitem[Wei et~al.(2020)Wei, Chen, Wei, Jiang, Chua, Zhou, and
  Jiang]{wei2020heuristic}
Zhipeng Wei, Jingjing Chen, Xingxing Wei, Linxi Jiang, Tat-Seng Chua, Fengfeng
  Zhou, and Yu-Gang Jiang.
\newblock Heuristic black-box adversarial attacks on video recognition models.
\newblock In \emph{Proceedings of the AAAI Conference on Artificial
  Intelligence}, pages 12338--12345, 2020.

\bibitem[Yan et~al.(2020)Yan, Wei, and Li]{yan2020sparse}
Huanqian Yan, Xingxing Wei, and Bo~Li.
\newblock Sparse black-box video attack with reinforcement learning.
\newblock \emph{arXiv preprint arXiv:2001.03754}, 2020.

\bibitem[Zhang et~al.(2020)Zhang, Zhu, Zhu, and Yang]{zhang2020motion}
Hu~Zhang, Linchao Zhu, Yi~Zhu, and Yi~Yang.
\newblock Motion-excited sampler: Video adversarial attack with sparked prior.
\newblock In \emph{European Conference on Computer Vision}. Springer, 2020.

\bibitem[Materzynska et~al.(2019)Materzynska, Berger, Bax, and
  Memisevic]{materzynska2019jester}
Joanna Materzynska, Guillaume Berger, Ingo Bax, and Roland Memisevic.
\newblock The jester dataset: A large-scale video dataset of human gestures.
\newblock In \emph{Proceedings of the IEEE/CVF International Conference on
  Computer Vision Workshops}, pages 0--0, 2019.

\bibitem[Liu et~al.(2017)Liu, Chen, Liu, and Song]{liu2016delving}
Yanpei Liu, Xinyun Chen, Chang Liu, and Dawn Song.
\newblock Delving into transferable adversarial examples and black-box attacks.
\newblock \emph{ICLR}, 2017.

\bibitem[Huang et~al.(2019)Huang, Katsman, He, Gu, Belongie, and
  Lim]{huang2019enhancing}
Qian Huang, Isay Katsman, Horace He, Zeqi Gu, Serge Belongie, and Ser-Nam Lim.
\newblock Enhancing adversarial example transferability with an intermediate
  level attack.
\newblock In \emph{Proceedings of the IEEE/CVF International Conference on
  Computer Vision}, pages 4733--4742, 2019.

\bibitem[Ilyas et~al.(2018{\natexlab{b}})Ilyas, Engstrom, Athalye, and
  Lin]{ilyas2018black}
Andrew Ilyas, Logan Engstrom, Anish Athalye, and Jessy Lin.
\newblock Black-box adversarial attacks with limited queries and information.
\newblock In \emph{International Conference on Machine Learning}, pages
  2137--2146. PMLR, 2018{\natexlab{b}}.

\bibitem[Kurakin et~al.(2017)Kurakin, Goodfellow, Bengio,
  et~al.]{kurakin2016adversarialphysical}
Alexey Kurakin, Ian Goodfellow, Samy Bengio, et~al.
\newblock Adversarial examples in the physical world, 2017.

\bibitem[Ren et~al.(2019)Ren, Zhao, and Ermon]{ren2019adaptive}
Hongyu Ren, Shengjia Zhao, and Stefano Ermon.
\newblock Adaptive antithetic sampling for variance reduction.
\newblock In \emph{International Conference on Machine Learning}, pages
  5420--5428. PMLR, 2019.

\bibitem[Hartley and Zisserman(2004)]{Hartley2004}
R.~I. Hartley and A.~Zisserman.
\newblock \emph{Multiple View Geometry in Computer Vision}.
\newblock Cambridge University Press, ISBN: 0521540518, second edition, 2004.

\bibitem[Soomro et~al.(2012)Soomro, Zamir, and Shah]{soomro2012ucf101}
Khurram Soomro, Amir~Roshan Zamir, and Mubarak Shah.
\newblock Ucf101: A dataset of 101 human actions classes from videos in the
  wild.
\newblock \emph{arXiv preprint arXiv:1212.0402}, 2012.

\bibitem[Tran et~al.(2015)Tran, Bourdev, Fergus, Torresani, and
  Paluri]{tran2015learning}
Du~Tran, Lubomir Bourdev, Rob Fergus, Lorenzo Torresani, and Manohar Paluri.
\newblock Learning spatiotemporal features with 3d convolutional networks.
\newblock In \emph{Proceedings of the IEEE international conference on computer
  vision}, pages 4489--4497, 2015.

\bibitem[Feichtenhofer et~al.(2019)Feichtenhofer, Fan, Malik, and
  He]{feichtenhofer2019slowfast}
Christoph Feichtenhofer, Haoqi Fan, Jitendra Malik, and Kaiming He.
\newblock Slowfast networks for video recognition.
\newblock In \emph{Proceedings of the IEEE/CVF International Conference on
  Computer Vision}, pages 6202--6211, 2019.

\bibitem[Yang et~al.(2020)Yang, Xu, Shi, Dai, and Zhou]{yang2020temporal}
Ceyuan Yang, Yinghao Xu, Jianping Shi, Bo~Dai, and Bolei Zhou.
\newblock Temporal pyramid network for action recognition.
\newblock In \emph{Proceedings of the IEEE/CVF Conference on Computer Vision
  and Pattern Recognition}, pages 591--600, 2020.

\bibitem[Carreira and Zisserman(2017)]{carreira2017quo}
Joao Carreira and Andrew Zisserman.
\newblock Quo vadis, action recognition? a new model and the kinetics dataset.
\newblock In \emph{proceedings of the IEEE Conference on Computer Vision and
  Pattern Recognition}, pages 6299--6308, 2017.

\bibitem[Moosavi-Dezfooli et~al.(2017)Moosavi-Dezfooli, Fawzi, Fawzi, and
  Frossard]{moosavi2017universal}
Seyed-Mohsen Moosavi-Dezfooli, Alhussein Fawzi, Omar Fawzi, and Pascal
  Frossard.
\newblock Universal adversarial perturbations.
\newblock In \emph{Proceedings of the IEEE conference on computer vision and
  pattern recognition}, pages 1765--1773, 2017.

\bibitem[Mopuri et~al.(2018)Mopuri, Ojha, Garg, and Babu]{mopuri2018nag}
Konda~Reddy Mopuri, Utkarsh Ojha, Utsav Garg, and R~Venkatesh Babu.
\newblock Nag: Network for adversary generation.
\newblock In \emph{Proceedings of the IEEE Conference on Computer Vision and
  Pattern Recognition}, pages 742--751, 2018.

\bibitem[Aich et~al.(2020)Aich, Gupta, Panda, Hyder, Asif, and
  Roy-Chowdhury]{aich2020non}
Abhishek Aich, Akash Gupta, Rameswar Panda, Rakib Hyder, M~Salman Asif, and
  Amit~K Roy-Chowdhury.
\newblock Non-adversarial video synthesis with learned priors.
\newblock In \emph{Proceedings of the IEEE/CVF Conference on Computer Vision
  and Pattern Recognition}, pages 6090--6099, 2020.

\bibitem[Gupta et~al.(2020)Gupta, Aich, and
  Roy-Chowdhury]{10.1145/3394171.3413686}
Akash Gupta, Abhishek Aich, and Amit~K. Roy-Chowdhury.
\newblock Alanet: Adaptive latent attention network for joint video deblurring
  and interpolation.
\newblock In \emph{Proceedings of the 28th ACM International Conference on
  Multimedia}, page 256–264, New York, NY, USA, 2020. Association for
  Computing Machinery.
\newblock ISBN 9781450379885.

\bibitem[Tulyakov et~al.(2018)Tulyakov, Liu, Yang, and
  Kautz]{Tulyakov:2018:MoCoGAN}
Sergey Tulyakov, Ming-Yu Liu, Xiaodong Yang, and Jan Kautz.
\newblock {MoCoGAN}: Decomposing motion and content for video generation.
\newblock In \emph{IEEE Conference on Computer Vision and Pattern Recognition
  (CVPR)}, pages 1526--1535, 2018.

\bibitem[Naseer et~al.(2019)Naseer, Khan, Khan, Khan, and
  Porikli]{naseer2019cross}
Muzammal Naseer, Salman~H Khan, Harris Khan, Fahad~Shahbaz Khan, and Fatih
  Porikli.
\newblock {Cross-Domain Transferability of Adversarial Perturbations}.
\newblock \emph{arXiv preprint arXiv:1905.11736}, 2019.

\bibitem[Poursaeed et~al.(2018)Poursaeed, Katsman, Gao, and
  Belongie]{poursaeed2018generative}
Omid Poursaeed, Isay Katsman, Bicheng Gao, and Serge Belongie.
\newblock {Generative Adversarial Perturbations}.
\newblock In \emph{Proceedings of the IEEE Conference on Computer Vision and
  Pattern Recognition}, pages 4422--4431, 2018.

\bibitem[Yin et~al.(2019)Yin, Kolouri, and Rohde]{yin2019gat}
Xuwang Yin, Soheil Kolouri, and Gustavo~K Rohde.
\newblock Gat: Generative adversarial training for adversarial example
  detection and robust classification.
\newblock In \emph{International Conference on Learning Representations}, 2019.

\bibitem[Wang et~al.(2019)Wang, Ma, Bailey, Yi, Zhou, and
  Gu]{wang2019convergence}
Yisen Wang, Xingjun Ma, James Bailey, Jinfeng Yi, Bowen Zhou, and Quanquan Gu.
\newblock On the convergence and robustness of adversarial training.
\newblock In \emph{ICML}, volume~1, page~2, 2019.

\bibitem[Li et~al.(2020)Li, Zhu, Paul, Roy-Chowdhury, Song, Krishnamurthy,
  Swami, and Chan]{li2020connecting}
Shasha Li, Shitong Zhu, Sudipta Paul, Amit Roy-Chowdhury, Chengyu Song,
  Srikanth Krishnamurthy, Ananthram Swami, and Kevin~S Chan.
\newblock Connecting the dots: Detecting adversarial perturbations using
  context inconsistency.
\newblock In \emph{European Conference on Computer Vision}, pages 396--413.
  Springer, 2020.

\bibitem[Wang et~al.(2021)Wang, Li, Liu, Wang, and
  Roy-Chowdhury]{wang2021multi}
Xueping Wang, Shasha Li, Min Liu, Yaonan Wang, and Amit~K Roy-Chowdhury.
\newblock Multi-expert adversarial attack detection in person re-identification
  using context inconsistency.
\newblock \emph{Proceedings of the IEEE/CVF International Conference on
  Computer Vision}, 2021.

\bibitem[Yin et~al.(2021)Yin, Li, Cai, Song, Asif, Roy-Chowdhury, and
  Krishnamurthy]{yin2021exploiting}
Mingjun Yin, Shasha Li, Zikui Cai, Chengyu Song, M~Salman Asif, Amit~K
  Roy-Chowdhury, and Srikanth~V Krishnamurthy.
\newblock Exploiting multi-object relationships for detecting adversarial
  attacks in complex scenes.
\newblock \emph{Proceedings of the IEEE/CVF International Conference on
  Computer Vision}, 2021.

\bibitem[Contributors(2020)]{2020mmaction2}
MMAction2 Contributors.
\newblock Openmmlab's next generation video understanding toolbox and
  benchmark.
\newblock \url{https://github.com/open-mmlab/mmaction2}, 2020.

\end{thebibliography}


\captionsetup[figure]{list=yes}
\captionsetup[table]{list=yes}
\newpage
\begin{center}
  \vspace*{2\baselineskip}
  \Large\bf{Adversarial Attacks on Black Box Video Classifiers:\\ Leveraging the Power of Geometric Transformations\\ (Supplementary Material)}  
\end{center}
\vspace{6\baselineskip}

{
\hypersetup{
    linkcolor=black
}
{\centering
 \begin{minipage}{\textwidth}
 \let\mtcontentsname\contentsname
 \renewcommand\contentsname{\MakeUppercase\mtcontentsname}
 \renewcommand*{\cftsecdotsep}{4.5}
 \noindent
 \rule{\textwidth}{1.4pt}\\[-0.75em]
 \noindent
 \rule{\textwidth}{0.4pt}
 \tableofcontents
 \rule{\textwidth}{0.4pt}\\[-0.70em]
 \noindent
 \rule{\textwidth}{1.4pt}
 \setlength{\cftfigindent}{0pt}
 \setlength{\cfttabindent}{0pt}
 \listoftables
 \listoffigures
 \end{minipage}\par}
}
\clearpage
\renewcommand\thesection{\Alph{section}}
\setcounter{section}{0}
\setcounter{figure}{0}
\setcounter{table}{0}
\resumetocwriting

\section{Victim Video Classifiers: Clean Test Accuracy}
\normalem
We consider four state-of-the-art video classification models, representing diverse methodologies of learning from videos, i.e., C3D \cite{tran2015learning}, SlowFast \cite{feichtenhofer2019slowfast}, TPN \cite{yang2020temporal} and I3D \cite{,carreira2017quo}, as our black-box victim models to perform adversarial attack. The \textit{C3D} model applies 3D convolution to learn spatio-temporal features from videos. \textit{SlowFast} uses a two-pathway architecture where the slow pathway operates at a low frame rate to capture spatial semantics and the fast pathway operates at a high frame rate to capture motion at fine temporal resolution. \textit{TPN}  captures actions at various tempos by using a feature-level temporal pyramid network. \textit{I3D} proposes the Inflated 3D ConvNet(I3D) with Inflated 2D filters and pooling kernels of traditional 2D CNNs. All the models are trained using open-source toolbox MMAction2 \cite{2020mmaction2} with their default setups. The test accuracy of the victim models with clean 16-frame videos on both UCF-101 and Jester datasets are shown in Table \ref{tab:recognition_accuracy}. Note that both datasets do not contain personally identifiable information and offensive contents.

\renewcommand{\arraystretch}{1.2}
\begin{table}[!ht]
\vspace*{-\baselineskip}
\centering
\caption{Clean test Accuracy of the victim classifiers}
\label{tab:recognition_accuracy}
\begin{tabular}{c|c|c|c|c}
\hline
\rowcolor{black!10}
 & \multicolumn{4}{c}{\textbf{Black-box Video Classifiers}}\\
\cline{2-5}
\rowcolor{black!10}
 \multirow{-2}{*}{\textbf{Datasets}} &  \multicolumn{1}{c|}{\textbf{C3D}}  & \multicolumn{1}{c|}{\textbf{SlowFast}} & \multicolumn{1}{c|}{\textbf{TPN}} & \multicolumn{1}{c}{\textbf{I3D}} \\
\hline
UCF-101 & 78.8\% & 85.4\% & 74.3\% & 71.7\% \\
\hline
Jester  & 90.1\% & 89.5\% & 90.5\% & 91.2\% \\
\hline
\end{tabular}
\end{table}

\section{Additional Experiments with Different Perturbation Budgets $\rho_{\max}$}
We present additional analysis of the attack performance of \geotrap and our two baseline methods, i.e., \HeuristicAttack and \MotionSamplerAttack for $\rho_{\max} = 8, 16$ in Table \ref{tab:budgets}. Note that for comprehensibility, we also provide the results for $\rho_{\max} = 10$ from the main manuscript in Table \ref{tab:budgets}. We observe that \geotrap consistently outperforms \MotionSamplerAttack; \geotrap requires less number of queries while achieves same or higher attack success rates.
\renewcommand{\arraystretch}{1.2}
\begin{table}[t]
\centering
\caption{Additional analysis of attack performance with different perturbation budgets $\rho_{\max}$}
\label{tab:budgets}
\resizebox{\columnwidth}{!}{
\begin{tabular}{c|c|cc|cc|cc|cc}
\hline
\rowcolor{black!10}
&  & \multicolumn{8}{c}{\textbf{Black-box Video Classifiers}}\\
\cline{3-10}
\rowcolor{black!10}
& & \multicolumn{2}{c|}{\textbf{C3D}}  & \multicolumn{2}{c|}{\textbf{SlowFast}}  & \multicolumn{2}{c|}{\textbf{TPN}} & \multicolumn{2}{c}{\textbf{I3D}} \\
\cline{3-10}
\rowcolor{black!10}
\multirow{-3}{*}{\textbf{Budget}} & \multirow{-3}{*}{\textbf{Methods}} & ANQ ($\downarrow$) & SR ($\uparrow$) & ANQ ($\downarrow$) & SR ($\uparrow$) & ANQ ($\downarrow$) & SR ($\uparrow$) & ANQ ($\downarrow$) & SR ($\uparrow$) \\ 
\hline
\rowcolor{yellow!10}
\multicolumn{10}{c}{\textbf{Attack: Untargeted, Dataset: Jester}} \\ \hline
\multirow{2}{*}{$\rho_{\max}=8$}  & \MotionSamplerAttack & 7310 & 96.3\% & 1926 & 100\% & 8056 & 91.3\% & 5482 & 98.1\% \\
\cline{2-10} 
& \geotrap (\textbf{Ours}) & \textbf{2614} & \textbf{100\%} & \textbf{553} & \textbf{100\%} & \textbf{4518} & \textbf{92.4\%} & \textbf{2312} & \textbf{100\%} \\ 
\hline
\multirow{2}{*}{$\rho_{\max}=10$} & \MotionSamplerAttack & 4549 & 99.0\% & 1906 & 100\% & 6269 & 91.3\% & 3029 & 99.4\% \\
\cline{2-10} 
& \geotrap (\textbf{Ours}) & \textbf{1602}         & \textbf{100\%}        & \textbf{521}          & \textbf{100\%}        & \textbf{3315}         & \textbf{92.4\%}       & \textbf{1599}         & \textbf{100\%}        \\ \hline
\multirow{2}{*}{$\rho_{\max}=16$} & \MotionSamplerAttack     & 2201                  & 100\%                 & 1421                  & 100\%                 & 3786                  & 96.3\%                & 1347                  & 100\%                 \\ \cline{2-10} 
& \geotrap (\textbf{Ours}) & \textbf{311}          & \textbf{100\%}        & \textbf{137}          & \textbf{100\%}        & \textbf{3147}         & \textbf{96.3\%}       & \textbf{551}          & \textbf{100\%}        \\ 
\hline
\rowcolor{yellow!10}
\multicolumn{10}{c}{\textbf{Attack: Untargeted, Dataset: UCF-101}}\\ 
\hline
\multirow{2}{*}{$\rho_{\max}=8$}  & \MotionSamplerAttack     & 16848          & 78.0\%          & 5436           & 95.0\%          & 20687           & 70.0\%          & 9242           & 92.0\%          \\ \cline{2-10} 
                                  & \geotrap (\textbf{Ours}) & \textbf{12100} & \textbf{84.0\%} & \textbf{2064}  & \textbf{98.0\%} & \textbf{18433}  & \textbf{74.0\%} & \textbf{6647}  & \textbf{97.0\%} \\ \hline
\multirow{2}{*}{$\rho_{\max}=10$} & \MotionSamplerAttack     & 14336          & 81.6\%          & 4673           & 97.2\%          & 20369           & 75.8\%          & 7400           & 94.4\%          \\ \cline{2-10} 
                                  & \geotrap (\textbf{Ours}) & \textbf{11490} & \textbf{86.2\%} & \textbf{1547}  & \textbf{98.8\%} & \textbf{17716}  & \textbf{76.1\%} & \textbf{4887}  & \textbf{97.4\%} \\ \hline
\multirow{2}{*}{$\rho_{\max}=16$} & \MotionSamplerAttack     & 11605          & 82.0\%          & 1944           & 99.\%           & 18055           & 75.8\%          & 4437           & 96.0\%          \\ \cline{2-10} 
                                  & \geotrap (\textbf{Ours}) & \textbf{9006}  & \textbf{86.2\%} & \textbf{858}   & \textbf{99.0\%} & \textbf{15972}  & \textbf{76.1\%} & \textbf{2643}  & \textbf{98.0\%} \\ \hline

\rowcolor{yellow!10}
\multicolumn{10}{c}{\textbf{Attack: Targeted, Dataset: Jester}}\\ 
\hline
\multirow{2}{*}{$\rho_{\max}=8$}  & \MotionSamplerAttack     & 42136                 & 92.6\%                & 39833                 & 98.1\%                & 121800                & 52.2\%                & 48788                 & 85.2\%                \\ \cline{2-10} 
& \geotrap (\textbf{Ours}) & \textbf{9333}         & \textbf{100\%}        & \textbf{11433}        & \textbf{98.1\%}       & \textbf{51799}        & \textbf{88.9\%}       & \textbf{25552}        & \textbf{96.3\%}       \\ 
\hline
\multirow{2}{*}{$\rho_{\max}=10$} & \MotionSamplerAttack     & 26704                 & 98.2\%                & 33087                 & 100\%                 & 63721                 & 80.9\%                & 39037                 & 90.7\%                \\ 
\cline{2-10} 
& \geotrap (\textbf{Ours}) & \textbf{6198}         & \textbf{100\%}        & \textbf{7788}         & \textbf{100\%}        & \textbf{41294}        & \textbf{92.6\%}       & \textbf{19542}        & \textbf{98.2\%}       \\ 
\hline
\multirow{2}{*}{$\rho_{\max}=16$} & \MotionSamplerAttack     & 8696                  & 100\%                 & 18901                 & 100\%                 & 40643                 & 90.7\%                & 25308                 & 94.4\%                \\ 
\cline{2-10} 
& \geotrap (\textbf{Ours}) & \textbf{4219}         & \textbf{100\%}        & \textbf{3855}         & \textbf{100\%}        & \textbf{16979}        & \textbf{96.3\%}       & \textbf{9110}         & \textbf{100\%}        \\ 
\hline
\rowcolor{yellow!10}
\multicolumn{10}{c}{\textbf{Attack: Targeted, Dataset: UCF-101}}\\ 
\hline
\multirow{2}{*}{$\rho_{\max}=8$}  & \MotionSamplerAttack     & 136327         & 51.7\%          & 72807          & 76.7\%          & 153355          & 35.0\%          & 107304         & 51.1\%          \\ \cline{2-10} 
                                  & \geotrap (\textbf{Ours}) & \textbf{90401} & \textbf{82.5\%} & \textbf{27306} & \textbf{93.0\%} & \textbf{150052} & \textbf{36.8\%} & \textbf{91773} & \textbf{59.3\%} \\ \hline
\multirow{2}{*}{$\rho_{\max}=10$} & \MotionSamplerAttack     & 100467                & 71.1\%                & 57126                 & 86.0\%                  & 151409                & 31.6\%                & 96498                 & 59.6\%                \\ 
\cline{2-10} 
& \geotrap (\textbf{Ours}) & \textbf{71820}        & \textbf{85.8\%}       & \textbf{21878}        & \textbf{95.0\%}         & \textbf{141629}       & \textbf{40.0\%}       & \textbf{76708}        & \textbf{74.6\%}       \\ 
\hline
\multirow{2}{*}{$\rho_{\max}=16$} & \MotionSamplerAttack     & 69344                 & 79.6\%                & 37759                 & 92.8\%                & 143504                & 45.0\%                & 70707                 & 75.0\%                \\ 
\cline{2-10} 
& \geotrap (\textbf{Ours}) & \textbf{35641}        & \textbf{98.0\%}       & \textbf{18177}        & \textbf{95.0\%}       & \textbf{132065}       & \textbf{45.5\%}       & \textbf{44400}        & \textbf{86.0\%}       \\ \hline
\end{tabular}
}
\vspace*{-\baselineskip}
\end{table}
\section{Statistical Comparison of Different Attack Methods}
\begin{table}[b]
\vspace*{-\baselineskip}
\centering
\caption{Statistical results with respect to the random seed after running attacks multiple times (\textit{Attack}: Targeted, \textit{victim classifier}: I3D, \textit{Dataset}: Jester, \textit{perturbation budget}: $\rho_{\max} = 16$)}
\label{tab:error_bars}
\resizebox{\columnwidth}{!}{
\begin{tabular}{c|cc|cc|cc}
\hline
\rowcolor{black!10}
 & \multicolumn{6}{c}{\textbf{Methods}}\\
\cline{2-7}
\rowcolor{black!10}
 & \multicolumn{2}{c|}{\textsc{Heuristic}}                      & \multicolumn{2}{c|}{\textsc{Motion Sampler}}                 & \multicolumn{2}{c}{\geotrap} \\ 
\cline{2-7}
\rowcolor{black!10}
 & ANQ ($\downarrow$) & SR ($\uparrow$) & ANQ ($\downarrow$) & SR ($\uparrow$) & ANQ ($\downarrow$) & SR ($\uparrow$) \\ 
\hline
\rowcolor{red!10}
Run 1 & 31088 & 77.9\% & 25308 & 94.4\% & 9110& 100\% \\
\rowcolor{red!10}
Run 2  & 38388  & 76.0\% & 20290 & 96.3\% &10110 & 100\%\\
\rowcolor{red!10}
Run 3  & 42098 & 74.1\% &  23356& 94.4\% & 5758& 100\%\\
\rowcolor{red!10}
Run 4  &  42022 & 74.0\% &  24464& 96.3\% & 7799& 100\%\\
\rowcolor{red!10}
Run 5  &  27431& 81.5\% &  25312& 94.4\% & 11782& 100\%\\
\hline
Mean  &  36205& 76.7\% & 23746 & 95.2\% & 8912& 100\%\\\hline
Standard Deviation  & 6643	 & 3.1\% & 2092 & 1.0\% & 2286& 0\%\\\hline
Standard Error  &2971  & 1.4\% & 936 & 0.5\% & 	1022 & 0\%\\
\hline
\end{tabular}
}
\end{table}
We have three sources of randomness in our experiments: \textit{a}) the sampling of $\bm{r}_{\text{frame}}$ in both \geotrap and \MotionSamplerAttack and the sampling of $\Phi_{\text{warp}}$ in \geotrap;  \textit{b}) direction initialization sampling in \HeuristicAttack; \textit{c}) target label sampling in targeted adversarial attacks for all three methods. To account for all these three randomness, we run the targeted attack against I3D model on Jester dataset under perturbation budget $\rho_{\max}=16$ for the three methods for five times. Using targeted attack strategy allows us to include the randomness of the target label sampling. We choose Jester dataset as it generally takes few queries to attack Jester dataset, thus saving testing time. We choose perturbation budget $\rho_{\max}=16$  as we observe that the attacks under such budget generally take few queries. We choose I3D model because compared to C3D and SlowFast, the attack success rates against I3D are not always 100\%; which is good for measuring the error bars for the attack success rates. In addition, compared to TPN, it generally takes fewer queries to launch the attack against I3D. We observe that the gradient estimated by \HeuristicAttack becomes zero after a certain number of iterations, in which case, no further queries are performed (and hence resulting in a low success rate).

\begin{figure}[t]
\centering
\captionsetup[subfigure]{labelformat=empty}
\hspace*{-1.5em}
\subfloat[]{
		\label{fig:error-anq}
		\includegraphics[width=0.49\columnwidth]{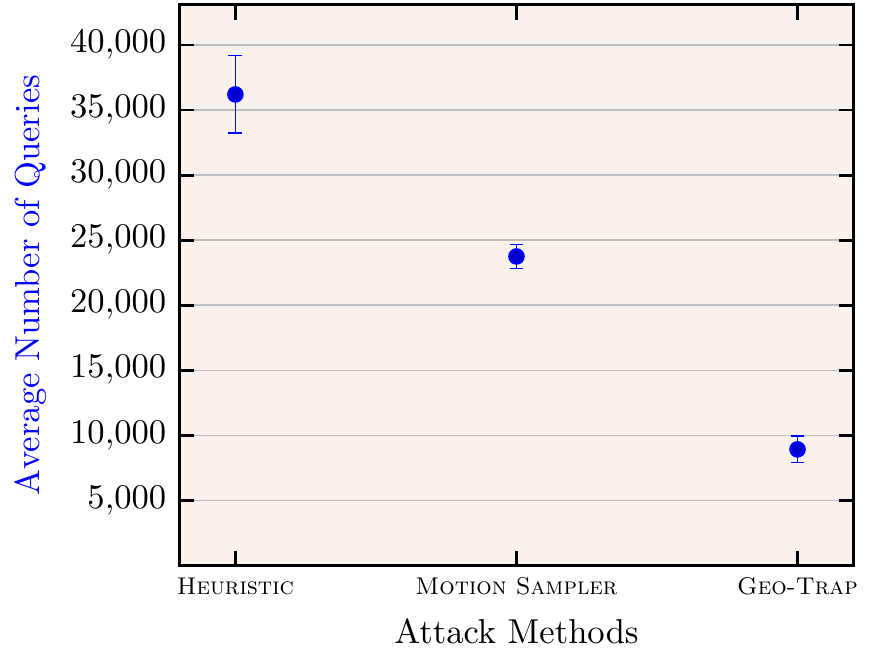} 
		} 
	\subfloat[]{
		\label{fig:error-sr}
		\includegraphics[width=0.475\columnwidth]{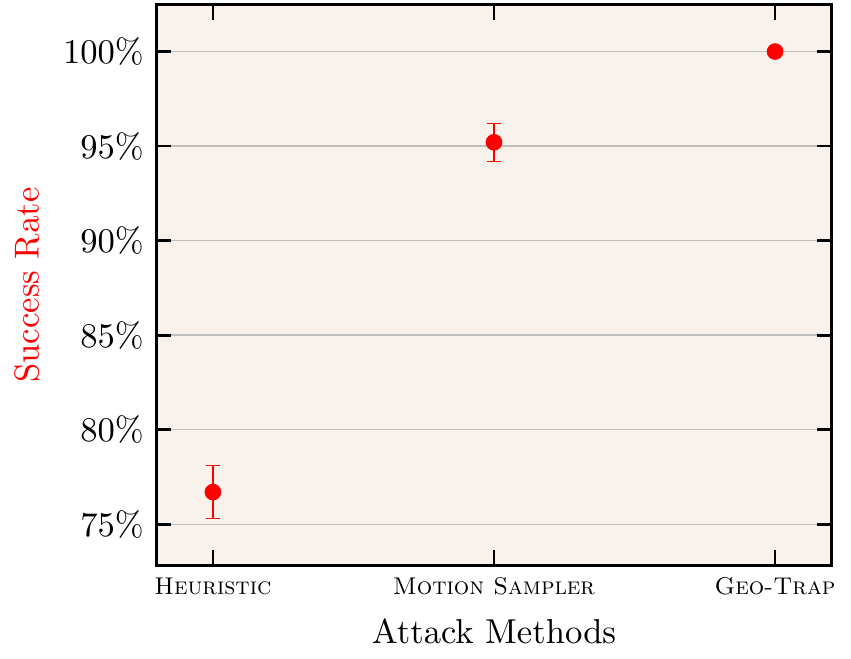} }
\vspace*{-\baselineskip}
\caption{Error bar plot to compare the performance (success rate and average number of queries) of different attack methods. We observe that our method outperforms the baseline methods in a statistically significant way. Detailed numbers are presented in Table \ref{tab:error_bars}}
\label{fig:error_bars}
\vspace*{-\baselineskip}
\end{figure}

We report the mean, standard deviation, and standard error in Table \ref{tab:error_bars} and present the error bar plot (with mean and standard error) in Figure \ref{fig:error_bars}. \geotrap, compared to other methods, requires statistically fewer number of queries while achieving statistically higher attack success rates than the baseline methods.
\section{Additional Experiments with Different Geometric Transformations}
\geotrap can employ different kinds of geometric transformations in the \textsc{Trans-Warp} function. In addition to the translation-dilation transformation ($D=3$) employed throughout the main manuscript, we report the performance of \geotrap with two other different geometric transformations, i.e., similarity transformation ($D=4$) and affine transformation ($D=6$). 

Recall that untargeted attack performance of \geotrap using these three geometric transformations on Jester dataset is reported in the main manuscript (\texttt{Figure 4}). In this section, we present the a more comprehensive set of results on both targeted and untargeted attacks, for both Jester and UCF-101 datasets in Table \ref{tab:dof}. We observe that the transformation with fewer degrees of freedom, i.e., translation-dilation transformation tends to requires fewer queries while having the same or higher attack success rates on Jester Dataset; this trend is consistent no matter which attack goal is used. On UCF-101 dataset, the transformations with fewer degrees of freedom, i.e., translation-dilation transformation and similarity transformation, require fewer queries while having the same or higher attack success rates compared to the affine transformation.

\renewcommand{\arraystretch}{1.2}
\begin{table}[t]
\centering
\caption{Additional analysis of attack performance of \geotrap with different geometric transformations $\mathcal{M}_{\bm{\phi}}$}
\label{tab:dof}
\resizebox{\columnwidth}{!}{
\begin{tabular}{c|cc|cc|cc|cc}
\hline
\rowcolor{black!10}
&  \multicolumn{8}{c}{\textbf{Black-box Video Classifiers}}\\
\cline{2-9}
\rowcolor{black!10}
 &  
\multicolumn{2}{c|}{\textbf{C3D}} & 
\multicolumn{2}{c|}{\textbf{SlowFast}} & 
\multicolumn{2}{c|}{\textbf{TPN}} & 
\multicolumn{2}{c}{\textbf{I3D}} \\ 
\cline{2-9}
\rowcolor{black!10}
\multirow{-3}{*}{\textbf{Geometric Transformations}, $\mathcal{M}_{\bm{\phi}}$} & ANQ ($\downarrow$) & SR ($\uparrow$) & ANQ ($\downarrow$) & SR ($\uparrow$) & ANQ ($\downarrow$) & SR ($\uparrow$) & ANQ ($\downarrow$) & SR ($\uparrow$) \\
\hline
\rowcolor{yellow!10}
\multicolumn{9}{c}{\textbf{Attack: Untargeted, Dataset: Jester}} \\ 
\hline
\textcolor{black}{Translation} & 3340 & 100\% & 1316 & 100\% & 5305 & 92.4\%  & 3943 & 100\% \\ \hline
\textcolor{black}{Dilation} & \textbf{1407} & \textbf{100\%} & \textbf{325} & \textbf{100\%} & {3574} & {92.4\%}  & \textbf{1239} & \textbf{100\%} \\ \hline
Translation Dilation & {1602} & {100\%} & {521} & {100\%} & \textbf{3315} & \textbf{92.4\%}  & {1599} & {100\%}\\ 
\hline
Similarity & 1621 & 100\% & 532 & 100\% & 3746 & 92.4\%  & 1629 & 100\% \\
\hline
Affine & 2716 & 100\% & 1057 & 100\% & 4579 & 91.6\%  & 2679 & 100\%\\
\hline
\rowcolor{yellow!10}
\multicolumn{9} {c}{\textbf{Attack: Targeted, Dataset: Jester}} \\ 
\hline
\textcolor{black}{Translation} & 12560 & 100\% & 18337 & 100\% &  56073& 83.0\%  & 46683 & 90.7\% \\ \hline
\textcolor{black}{Dilation} & 6887 & 100\% & 8134 & 98.1\% &  \textbf{36898}& \textbf{92.6}\%  & \textbf{14019} & \textbf{98.2\%} \\ \hline
Translation Dilation & \textbf{6198} & \textbf{100\%} & \textbf{7788} & \textbf{100\%} & {41294} & {92.6\%}  & {19542} & {98.2\%}\\ 
\hline
Similarity & 6431 & 100\% & 7939 & 100\% & 42594 & 90.7\%  & 19369 & 98.2\% \\
\hline
Affine & 10326 & 100\% & 15360 & 100\% & 55276 & 90.7\%  & 32006& 94.4\%\\
\hline
\rowcolor{yellow!10}
\multicolumn{9} {c}{\textbf{Attack: Untargeted, Dataset: UCF-101}} \\ \hline
\textcolor{black}{Translation} & 13145 & 86.2\% & 3959 & 98.0\% & 18551 & 3220\%  &{9078} & {94.0}\%\\  \hline
\textcolor{black}{Dilation} & \textbf{9991} & \textbf{87.6\%} & {1510} & \textbf{98.9\%} & \textbf{16847} & \textbf{76.7\%}  & \textbf{3755} & \textbf{97.4\%} \\ \hline
Translation Dilation & {11490} &{86.2\%} & {1547} & {98.9}\% & {17716} & 76.1\%  & {4887} & {97.4}\%\\ 
\hline
Similarity & {10624} & 85.8\% & \textbf{1489} & 98.6\% & {17492} & {76.7}\%  & 5694 & 95.0\% \\
\hline
Affine & 12792 & 84.8\% &  3088& 98.0\% & 17773 & 75.0\%  & 8291 & 94.0\%\\
\hline
\end{tabular}
}
\vspace*{-\baselineskip}
\end{table}
 
\section{Additional Experiments on \geotrap with Different Loss Functions}

\begin{figure}[htbp]
	\centering
	\subfloat[Jester dataset]{
		\label{fig:jes-prod}
		\includegraphics[width=0.49\columnwidth]{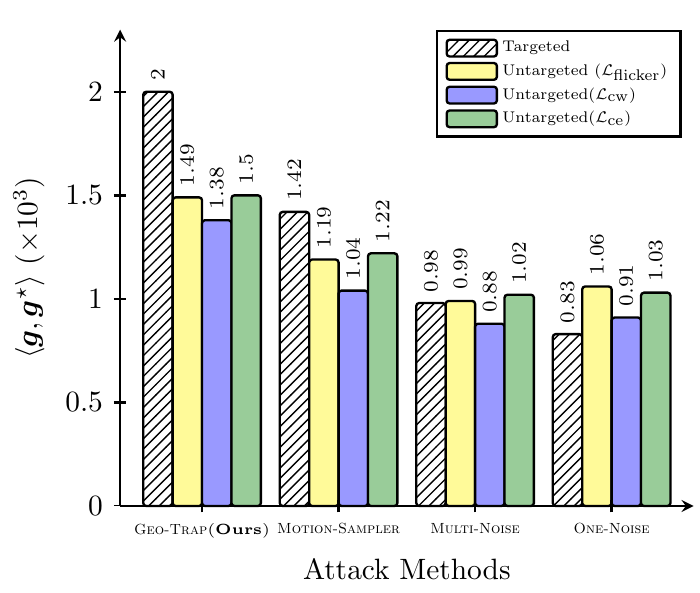} 
		} 
	\subfloat[UCF-101 dataset]{
		\label{fig:ucf-prod}
		\includegraphics[width=0.485\columnwidth]{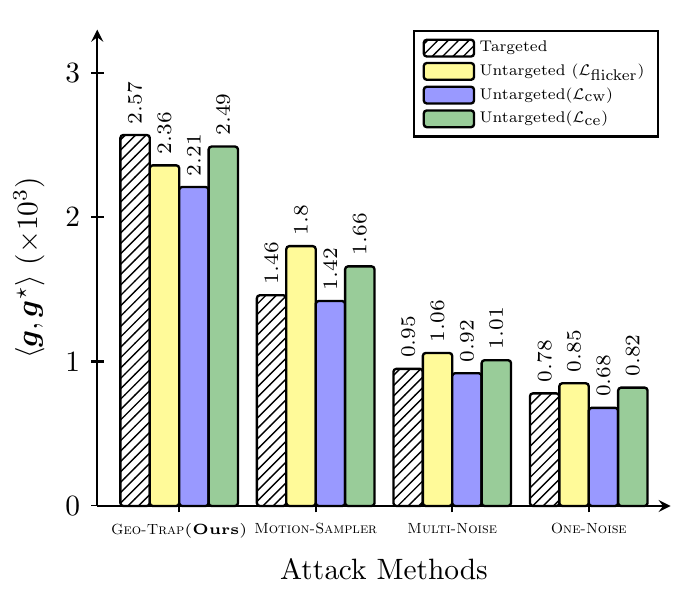} }
 	\caption{Evaluation of gradient estimation quality by calculating the cosine similarity between the ground truth gradient $\bm{g}^\star$ and the estimated gradient $\bm{g}$ calculated by different attack methods.}
	\label{fig:prod-supp} 
\end{figure}

In this section, we further validate that, compared to our three baseline methods (i.e., \MultiNoiseAttack, \OneNoiseAttack, \MotionSamplerAttack), the gradients searched with \geotrap are better. This is demonstrated by the fact that \geotrap's gradients generally have larger cosine similarity with the ground truth gradients. This trend is loss function agnostic, with both untargeted and targeted attacks, as shown in Figure \ref{fig:prod-supp}. We consider four attack loss functions, three untargeted attack loss functions and one targeted attack loss function, described below. 

We start with explaining the flicker loss used for untargeted attack and the cross-entropy loss used for targeted attack in the main paper.
Flicker loss is defined with the probability scores of the top-$2$ labels returned by $\bm{f}_{\bm{\theta}}(\bm{x})$ following \cite{pony2020over}.
In particular, if the attack is not successful, the most likely label predicted by $\bm{f}_{\bm{\theta}}(\bm{x})$ will be the true label $y$. We denote the probability score associated with this label as $p_y(\bm{x})$. Similarly, we denote the \textit{second} most likely label predicted by $\bm{f}_{\bm{\theta}}(\bm{x})$ as $y'$ and its corresponding probability score as $p_{y'}(\bm{x})$.  The loss function is defined to encourage $p_{y'}(\bm{x})$ increasing and  $p_y(\bm{x})$ decreasing until  $p_{y'}(\bm{x}) > p_y(\bm{x})$ and $y'$ becomes the predicted top-1 label. This loss function can be mathematically denoted as follows. 
\begin{equation}
    \mathcal{L}_{\text{flicker}}(\bm{x},y) = \bigg{[} \operatorname{min}\bigg{(}\dfrac{1}{m}\mathcal{K}(\bm{x},y)^2, \mathcal{K}(\bm{x},y)\bigg{)}\bigg{]}_+
    \text{with, } \mathcal{K}(\bm{x},y) = p_y(\bm{x}) - {p_{y'}(\bm{x})} + m
    \label{eq:loss-flicker}
\end{equation}
Here, $[a]_+ = \max(0, a)$ and $m>0$ is the desired margin of the original class probability below the adversarial class probability. We refer readers to \cite{pony2020over} for more detailed explanation of \eqref{eq:loss-flicker}. 

For the targeted attack, the cross-entropy loss is defined as follows.
\begin{equation}
\mathcal{L}(\bm{x},y_{\top}) = - \operatorname{log}\big{(} p_{y_{\top}}(\bm{x}) \big{)}
\label{eq:tar-loss}
\end{equation}
where $p_{y_{\top}}(\bm{x})$ is the probability score of the target label returned by $\bm{f}_{\bm{\theta}}(\bm{x})$.  

In addition to the above loss functions, we consider two other untargeted loss functions for gradient analysis of attacks methods. The first one is the untargeted attack loss function defined in \cite{zhang2020motion} based on CW2 loss \cite{carlini2017towards} as shown in the following.
\begin{equation}
    \mathcal{L}_{\text{cw}}(\bm{x},y) = \big{[} p_y(\bm{x}) - {p_{y'}(\bm{x})}\big{]}_+
\end{equation}
where, $p_y(\bm{x})$ is the largest probability score, which should be associated with the true label $y$, and  $p_{y'}(\bm{x})$ is the second largest probability score, which is associated with the second most confident label $y'$. The second loss is a cross-entropy loss where a lower $p_y(\bm{x})$ is encouraged, as shown in the following.
\begin{equation}
\mathcal{L}_{\text{ce}}(\bm{x},y) = - \operatorname{log}\big{(} 1- p_{y}(\bm{x}) \big{)}
\end{equation}

We calculate the average cosine similarity (over 1000 randomly chosen samples) between the ground truth gradients and the estimated gradients for \geotrap and the three baselines. As shown in Figure \ref{fig:prod-supp}, for all the five different loss functions considered and on both Jester (see Figure \ref{fig:jes-prod}) and UCF-101 (see Figure \ref{fig:ucf-prod}) dataset, the gradients searched by \geotrap  have better quality consistently. This explains why \geotrap requires less number of queries while achieving the same or higher attack success rates.

\section{Additional Examples of Adversarial Videos}
In this section, we provide additional adversarial examples on both Jester and UCF-101 datasets as shown in Figure \ref{fig:visualization_supp}. We observe that the generated adversarial frames have little difference from the clean ones but can lead to a failed classification. 
\begin{figure}[htbp]
	\centering
	\captionsetup[subfigure]{labelformat=empty}
	\subfloat[]{
		\includegraphics[height =0.45\textheight, width=\columnwidth]{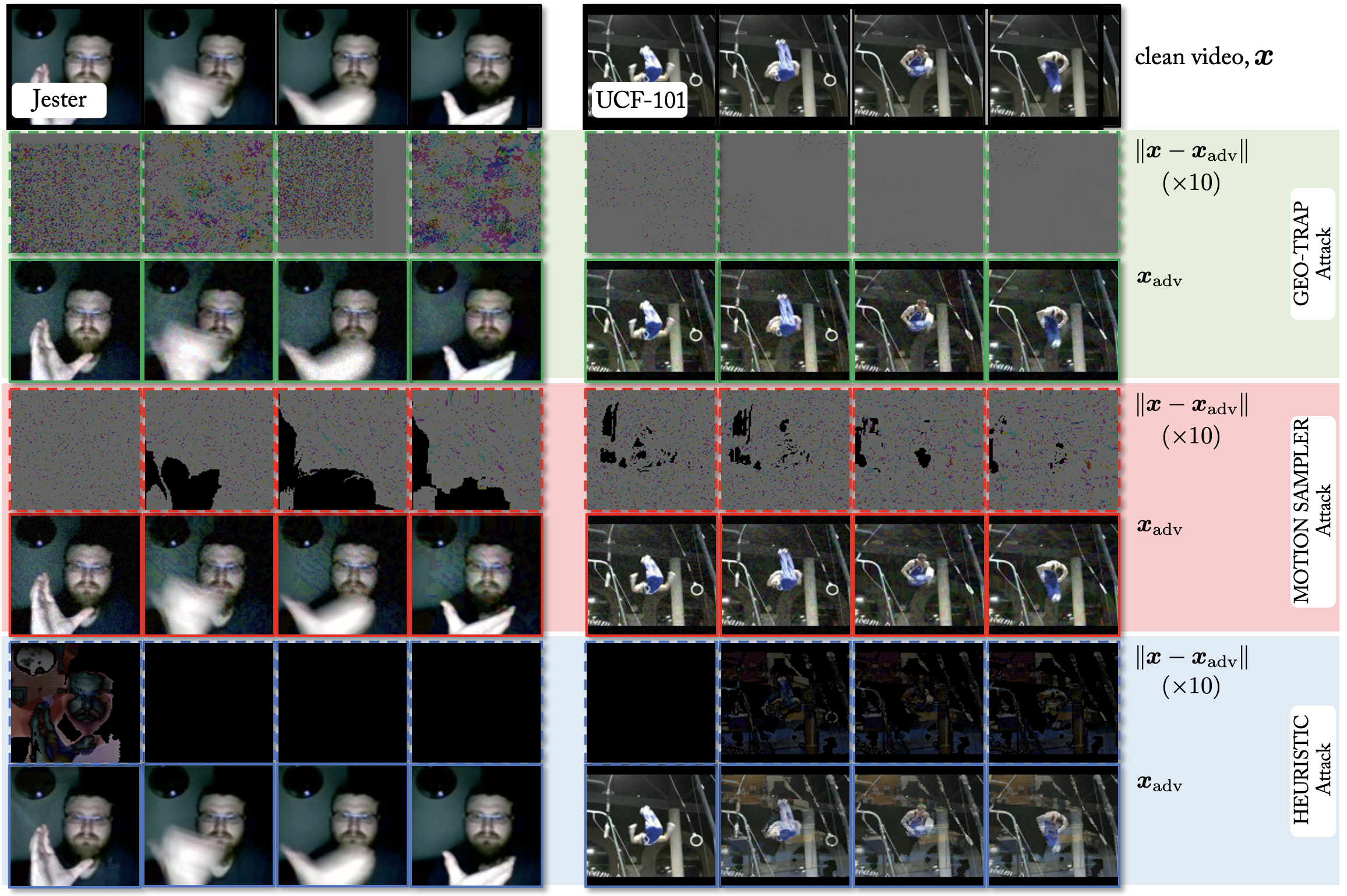} }
	\\	
	\subfloat[]{
		\includegraphics[height =0.45\textheight, width=\columnwidth]{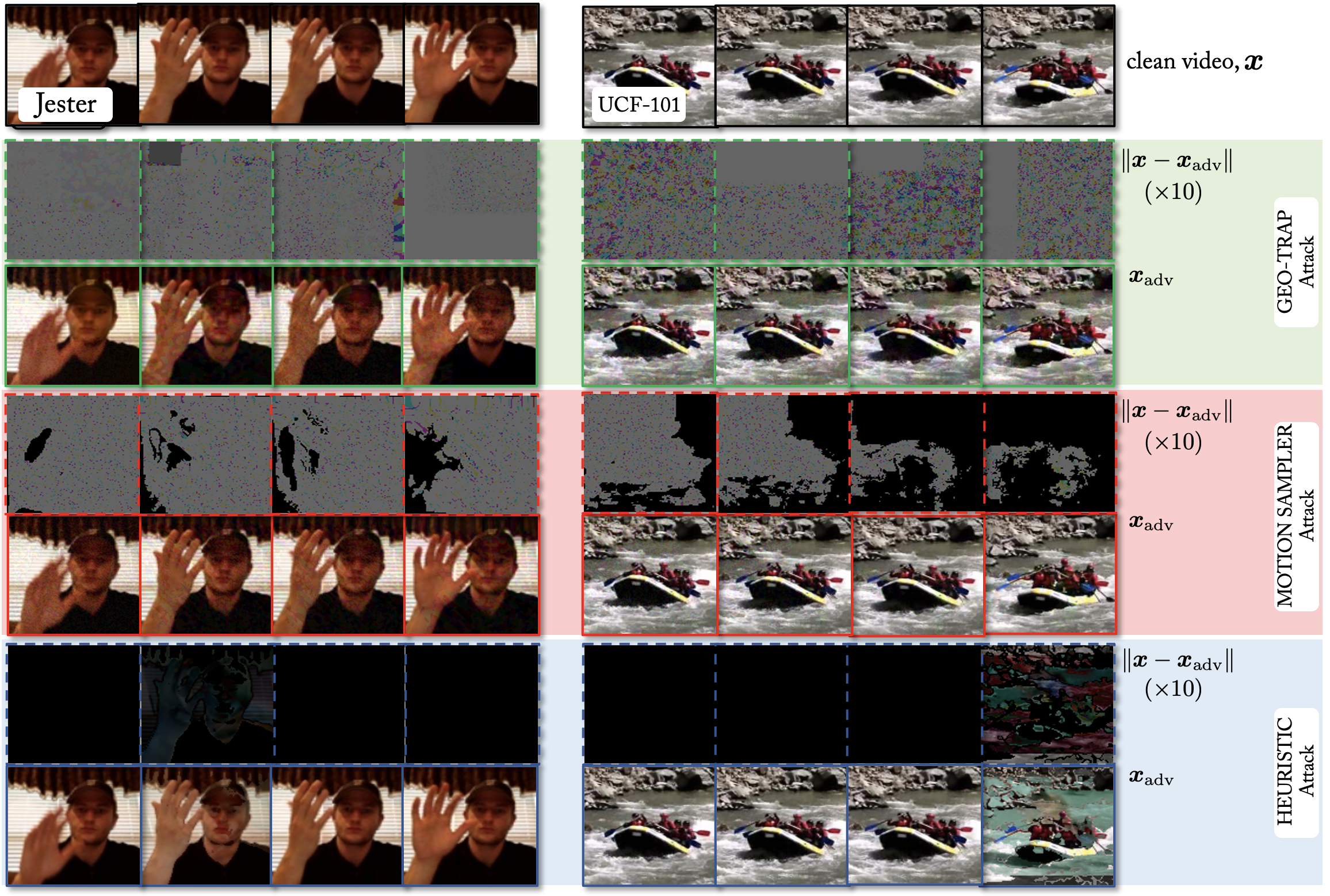} }
\end{figure}

\begin{figure}[htbp]
	\centering
    \ContinuedFloat
    \captionsetup[subfigure]{labelformat=empty}
	\subfloat[]{
		\includegraphics[height =0.45\textheight, width=\columnwidth]{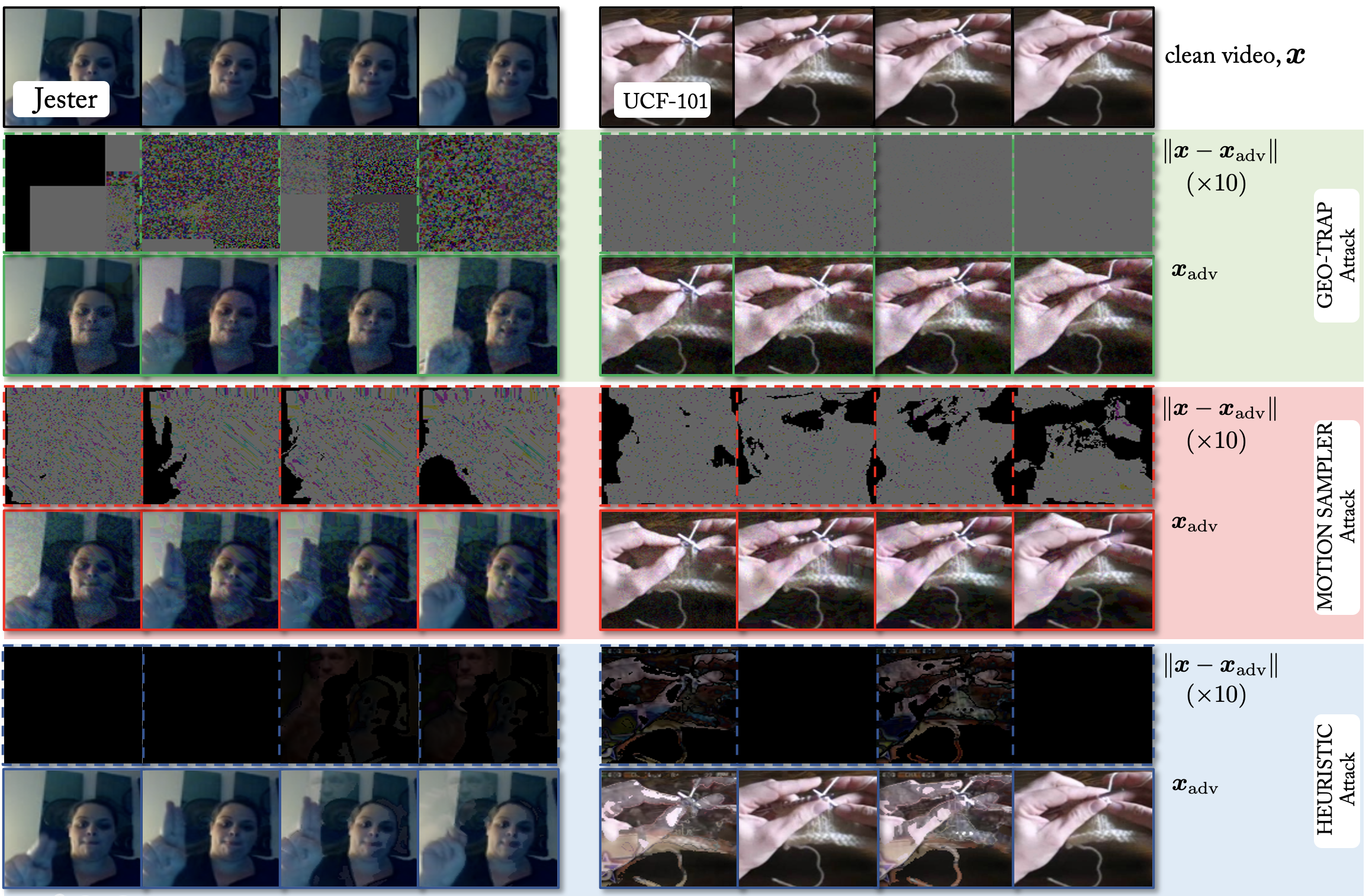} } 
	\\
	\subfloat[]{
		\includegraphics[height =0.45\textheight, width=\columnwidth]{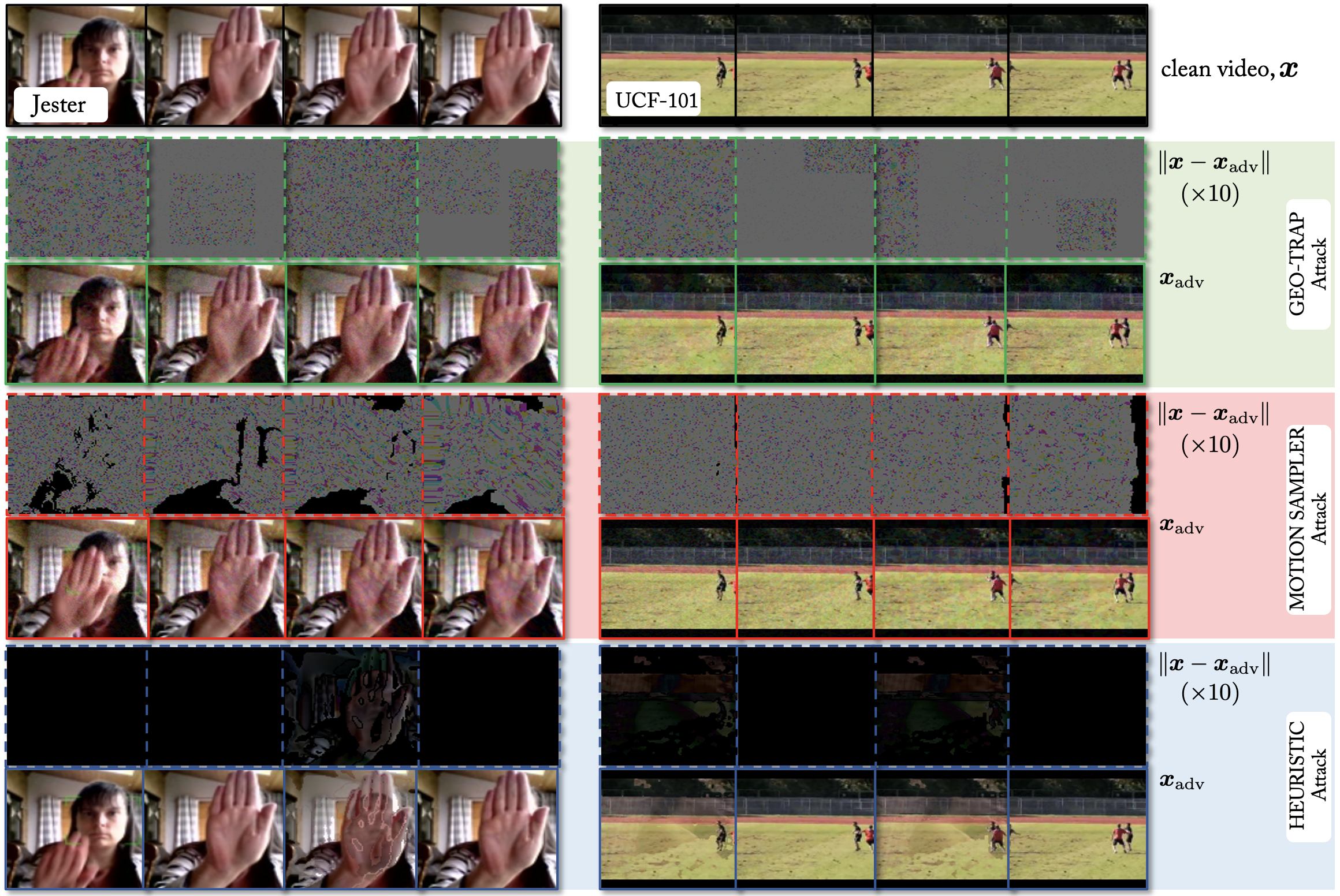} } 
 	\caption{The visualization of the perturbation ($\times 10$) and adversarial frames of our methods and the two baseline methods on Jester (left column) and UCF-101 datasets (right column).}
	\label{fig:visualization_supp} 
\end{figure}

\color{black}{In addition, we calculate PSNR to measure the perception of perturbations. We measure the minimum PSNR among all frames as it represents the worst-case scenario of maximum degradation for the video. For this, we generate the adversarial examples for untargeted attack against the C3D model on the Jester dataset. The average minimum (across all videos) PSNR of resultant adversarial videos for GEO-TRAP is 28.30 dB; for \MotionSamplerAttack is 28.60 dB, and for \HeuristicAttack is 22.06 dB. We observe that GEO-TRAP, as well as MotionSampler, has less video quality degradation compared to HeuristicAttack. }

\end{document}